\documentclass[10pt,twocolumn,letterpaper]{article}

\usepackage[pagenumbers]{cvpr}              

\usepackage{wrapfig}
\definecolor{green_zjr}{RGB}{6,104,86}
\usepackage[dvipsnames]{xcolor}
\usepackage{pifont}  

\newcommand{\et}[2]{${#1}^{\pm{#2}}$}
\newcommand{\etb}[2]{$\mathbf{{#1}}^{\pm{#2}}$}

\definecolor{LightGray}{rgb}{0.9,0.9,0.9}
\definecolor{DarkRed}{rgb}{0.75,0,0}
\definecolor{DarkBlue}{rgb}{0,0,0.55}
\definecolor{DarkGreen}{rgb}{0.43, 0.68, 0.28}
\usepackage{colortbl}
\usepackage{algorithm}
\usepackage{algorithmic}
\usepackage{amsmath}

\usepackage{multirow}

\usepackage{amsmath,amsfonts,bm}

\def\eqref#1{equation~\ref{#1}}

\def\1{\bm{1}}

\def\vc{{\bm{c}}}

\def\vm{{\bm{m}}}

\def\vx{{\bm{x}}}

\def\vz{{\bm{z}}}

\def\mC{{\bm{C}}}

\DeclareMathAlphabet{\mathsfit}{\encodingdefault}{\sfdefault}{m}{sl}
\SetMathAlphabet{\mathsfit}{bold}{\encodingdefault}{\sfdefault}{bx}{n}

\definecolor{cvprblue}{rgb}{0.21,0.49,0.74}
\usepackage[pagebackref,breaklinks,colorlinks,citecolor=cvprblue]{hyperref}

\def\paperID{***} 
\def\confName{CVPR}
\def\confYear{2026}

\title{\textsc{DeMoGen}: Towards Decompositional Human Motion Generation \\ with Energy-Based Diffusion Models}

\author{Jianrong Zhang$^{1}$, 
Hehe Fan$^{2,\dagger}$, 
Yi Yang$^2$ \\
\small \qquad $^{\dagger}$Corresponding author \\
\small $^1$ReLER, AAII, University of Technology Sydney \qquad
$^2$CCAI, Zhejiang University \\
\small \url{https://jiro-zhang.github.io/DeMoGen/}
}

\begin{document}
\maketitle
\begin{abstract}

Human motions are compositional: complex behaviors can be described as combinations of simpler primitives. However, existing approaches primarily focus on forward modeling, \eg, learning holistic mappings from text to motion or composing a complex motion from a set of motion concepts.
In this paper, we consider the inverse perspective: decomposing a holistic motion into semantically meaningful sub-components. We propose \textsc{DeMoGen}, a compositional training paradigm for decompositional learning that employs an energy-based diffusion model. This energy formulation directly captures the composed distribution of multiple motion concepts, enabling the model to discover them without relying on ground-truth motions for individual concepts.
Within this paradigm, we introduce three training variants to encourage a decompositional understanding of motion: \ding{182} \textsc{DeMoGen-Exp} explicitly trains on decomposed text prompts; \ding{183} \textsc{DeMoGen-OSS} performs orthogonal self-supervised decomposition; \ding{184} \textsc{DeMoGen-SC} enforces semantic consistency between original and decomposed text embeddings.
These variants enable our approach to disentangle reusable motion primitives from complex motion sequences. We also demonstrate that the decomposed motion concepts can be flexibly recombined to generate diverse and novel motions, generalizing beyond the training distribution. 
Additionally, we construct a text-decomposed dataset to support compositional training, serving as an extended resource to facilitate text-to-motion generation and motion composition. 
\end{abstract}

\section{Introduction}
\label{sec:intro}
\vspace{-1mm}
Humans are inherently capable of decomposing complex motions into a set of simpler motion primitives. For example, we can easily interpret a motion ``walking in a zigzag pattern while waving left hand'' as a composition of two motion concepts, \ie, ``walking in a zigzag pattern'' and ``waving left hand''. Such decompositional understanding allows humans to perform unusual motions by recombining known primitives, without extensive practice or prior exposure.

\begin{figure*}[t]
\centering
\includegraphics[width=\textwidth]{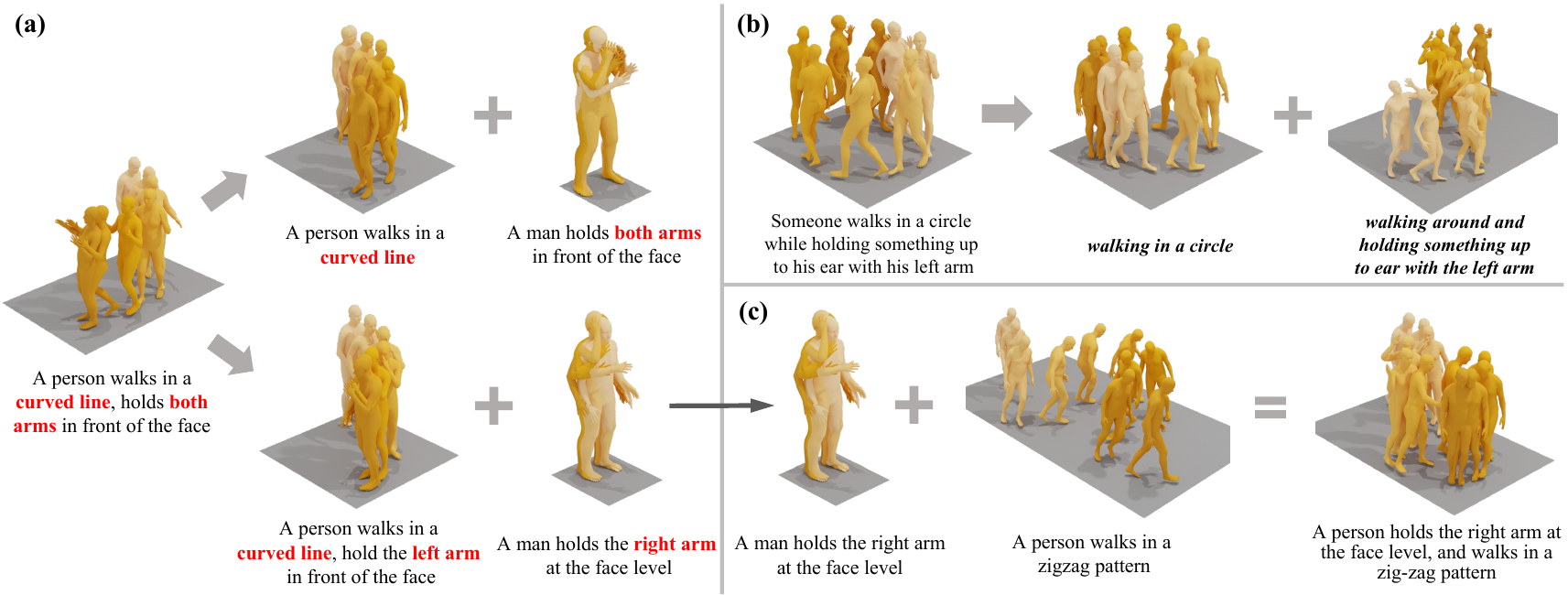}
\vspace{-7mm}
\caption{\textbf{Decompositional motion generation.} (a) Our approach is able to decompose a holistic motion into multiple motion concepts with \textsc{DeMoGen-Exp}.
This process is explicitly guided by decompositional textual descriptions, yielding diverse decomposition outcomes when varying the text cues. (b) Our approach also supports unguided decomposition using \textsc{DeMoGen-OSS} and \textsc{DeMoGen-SC}, where the model infers motion concepts without explicit decompositional text. We manually caption the inferred concepts in italic for easy understanding. (c) The decomposed motion primitives can be recombined to synthesize diverse and novel motions.
}
\vspace{-4mm}
\label{fig:teaser}
\end{figure*} 

In the past few years, text-to-motion generation has achieved remarkable progress, mainly attributed to the availability of large-scale motion datasets and advances in powerful generative models. Most state-of-the-art approaches focus on producing high-fidelity motions that are semantically aligned with a single textual description~\cite{Zhang_2023_CVPR,tevet2022MDM,zhang2022motiondiffuse,chen2022mld,zhang2023remodiffuse,attt2m,petrovich22temos,guo2024momask}. In parallel, several works have begun to explore compositional motion generation~\cite{TEACH:3DV:2022,petrovich24stmc,SINC:ICCV:2023,xie2023omnicontrol,karunratanakul2023guided,shafir2024priormdm}. They leverage pre-trained diffusion models through spatial editing and temporal concatenation for fine-grained control and generating long-duration motions, respectively. More recently, EnergyMoGen~\cite{zhang2025energymogen} presents two spectrums of energy-based models and achieves motion composition through energy aggregation. The approach also introduces additional logical operators, \ie, negation and combination of conjunction and negation, to enable more diverse forms of composition.

Despite these advances, existing approaches treat motion as a holistic sequence, without capturing its structured composition of motion concepts. This lack of decompositional understanding limits their ability to infer and recombine concepts like humans do. A natural question thus arises: \textit{how can we endow generative models with a human-like ability to disentangle the underlying motion concepts?}

As a response to this question, we propose \textsc{DeMoGen}, a compositional training paradigm that learns to decompose motions into a set of semantically interpretable motion concepts under compositional supervision.
Building upon the intrinsic relationship between diffusion models~\cite{sohl2015deep,ho2020denoising,rombach2022high} and Energy-Based Models(EBMs)~\cite{lecun2006tutorial,zhao2017energybased,du2019implicit}, we represent each motion concept as an energy score. Unlike previous methods that typically learn a one-to-one mapping between text and motion, our approach models the composed energy distribution of multiple motion concepts during training. This design enforces concept-level factorization, which naturally enables motion decomposition at inference, and further supports flexible concepts recombination for compositional motion generation (Figure~\ref{fig:teaser}(a), (b), (c)).

More specifically, we explore three variants of \textsc{DeMoGen}, reflecting varying degrees of compositional supervision: 
\ding{182} \textsc{DeMoGen-Exp} introduces \textit{\textbf{explicit supervision}} by training on decomposed text prompts, where each motion concept is conditioned on its corresponding textual description. Such explicit pairing enforces clear semantic boundaries for precise and diverse concept factorization (Figure~\ref{fig:teaser}(a));
\ding{183} \textsc{DeMoGen-OSS} operates under \textit{\textbf{orthogonal self-supervision}}. We partition the holistic text embedding into multiple segments and directly train the model on these segments. An orthogonality loss is applied to encourage disentanglement without additional explicit constraints, enabling unguided decomposition and improving motion diversity.
\ding{184} \textsc{DeMoGen-SC} leverages \textit{\textbf{semantic consistency supervision}} by aligning each partitioned text embedding segment with the corresponding decomposed text embeddings. This also enables unguided decomposition at inference (Figure~\ref{fig:teaser}(b)).

To support explicit and semantic consistency supervisions, we construct a DecompML dataset, an extension of the widely used HumanML3D~\cite{guo2022generating} with \textit{decompositional} text annotations. We break down original holistic motion descriptions into semantically interpretable concepts using large language models. Importantly, our dataset can also serve as a benchmark for compositional motion generation, promoting future research in this field.

\textsc{DeMoGen} is a unified training paradigm that simultaneously facilitates text-to-motion generation, multi-concept generation, as well as decompositional and compositional (including concept recombination) motion generation. With its general formulation, our approach is compatible with both latent and semantic-aware EMBs~\cite{zhang2025energymogen} and
can seamlessly integrate with mainstream diffusion-based methods. Extensive experiments demonstrate the effectiveness of \textsc{DeMoGen}, showing robust improvements on both text-to-motion and motion composition benchmarks \ie, HumanML3D~\cite{guo2022generating} and MTT~\cite{petrovich24stmc}. 
In particular, we observe significant improvements in the compositional and multi-concept motion generation tasks. Moreover, by pairing decomposed motions with their corresponding text concepts, DecompML can serve as an augmented motion-language dataset, which further benefits text-to-motion generation.
\section{Related Works}
\noindent\textbf{Text-to-Motion Generation.}
Text-to-motion generation, which seeks to generate temporally coherent and physically plausible motion sequences from text instructions, has attracted increasing attention in recent years. Early studies in this area are typically based on autoencoder architectures~\cite{ahuja2019language2pose,ghosh2021synthesis,tevet2022motionclip}. They establish cross-modal correspondences by projecting text and motion into a shared latent representation, but suffer from deterministic mappings that limit motion diversity and realism. Researchers~\cite{guo2022generating,petrovich22temos,petrovich21actor,petrovich2023tmr} adopt probabilistic frameworks, \ie, Variational Autoencoders (VAEs), which shift the focus from fixed encodings to distribution-based modeling, allowing diverse motion generation through sampling. Recent works introduce both autoregressive~\cite{Zhang_2023_CVPR,attt2m,humantomato,chuan2022tm2t,jiang2023motiongpt,zhang2023motiongpt} and non-autoregressive~\cite{pinyoanuntapong2024mmm,guo2024momask} strategies inspired by natural language~\cite{brown2020gpt3} and vision tasks~\cite{chang2022maskgit}, where the motion is represented as a sequence of discrete tokens via Vector Quantized-Variational Autoencoders (VQ-VAEs). For example, T2M-GPT~\cite{Zhang_2023_CVPR} adopts a GPT-like architecture, while MoMask~\cite{guo2024momask} applies masked token prediction to generate motion from text prompts.

In parallel, diffusion models have emerged as a powerful alternative for motion synthesis. These models leverage iterative refinement to capture complex spatial-temporal dynamics. MotionDiffuse~\cite{zhang2022motiondiffuse} and MDM~\cite{tevet2022MDM} first show the feasibility of diffusion on text-to-motion generation. This line of research~\cite{zhang2023remodiffuse,zhang2023finemogen,dabral2022mofusion,goel2024iterative,han2024amd,ren2023realistic,jin2024act,wang2023fg,xie2024towards,yang2023synthesizing,zhou2023ude,zhai2023language,yuan2023physdiff} applies diffusion directly to raw motion representations such as joint positions and rotations. 
Several works~\cite{chen2022mld,kong2023priority,lou2023diversemotion,gao2024guess} continue to investigate latent diffusion models~\cite{rombach2022high}. MLD~\cite{chen2022mld} is a representative motion latent diffusion model, and subsequent works~\cite{kong2023priority,lou2023diversemotion,gao2024guess} introduce architectural improvements, including GNN-based VAEs~\cite{hong2025salad} and the Mamba framework~\cite{zhang2025motion,gu2023mamba}. 

In contrast to these methods that learn a holistic text-to-motion mapping, we focus on a decompositional understanding of motion using an energy-based diffusion model. We demonstrate how a complex motion can be decomposed into a set of semantically interpretable motion concepts, and further show that these concepts can be flexibly recomposed to synthesize novel motions unseen during training.

\vspace{1.0mm}
\noindent\textbf{Compositional and Decompositional Generation.} Compositional and decompositional generation play a crucial role in achieving controllability and interpretability in generative modeling, and have been well explored in the image domain. 
Some compositional image generation approaches incorporate additional training strategies, such as contrastive learning~\cite{li2022stylet2i,cong2023attribute} and classifier guidance~\cite{shi2023exploring,garipov2023compositional}. Another line of work achieves compositional control at inference by modifying attention maps~\cite{feng2023trainingfree,park2024energy,hu2024statistical,hu2024provably} or composing the distributions of pre-trained generative models~\cite{du2021unsupervised,du2019implicit,liu2021learning,liu2022compositional,du2024compositional,du2020compositional}. As for the decompositional generation, previous studies~\cite{shen2020interpreting,harkonen2020ganspace,peebles2020hessian,preechakul2022diffusion,singh2019finegan} aim to disentangle desired factor by operating in the latent space. Recently, inspired by EBMs, Liu \etal~\cite{liu2023unsupervised} leveraged pre-trained diffusion models to discover diverse visual concepts (\eg, image class and style). Su \etal~\cite{su2024compositional} focused on compositional concept decomposition, enabling both global and local concept factorization from a given image. In the compositional motion generation domain~\cite{TEACH:3DV:2022,SINC:ICCV:2023,pinyoanuntapong2024controlmm}, most diffusion-based solutions~\cite{shafir2024priormdm,karunratanakul2023gmd,xie2023omnicontrol} incorporate spatial and temporal control into the denoising process based on MDM~\cite{tevet2022MDM} architecture. EnergyMoGen~\cite{zhang2025energymogen} formulates the denoising network and cross-attention as energy functions and combines their energy distributions in the latent space.

In this paper, we propose a compositional training paradigm for motion decomposition. We investigate three supervision regimes, ranging from explicit supervision to a fully self-supervised setting. By leveraging this paradigm, our approach seamlessly supports text-to-motion generation, motion composition, and motion decomposition within a single model.

\section{Method}
\label{sec:method}
\begin{figure*}[tp]
\centering
\includegraphics[width=1.0\textwidth]{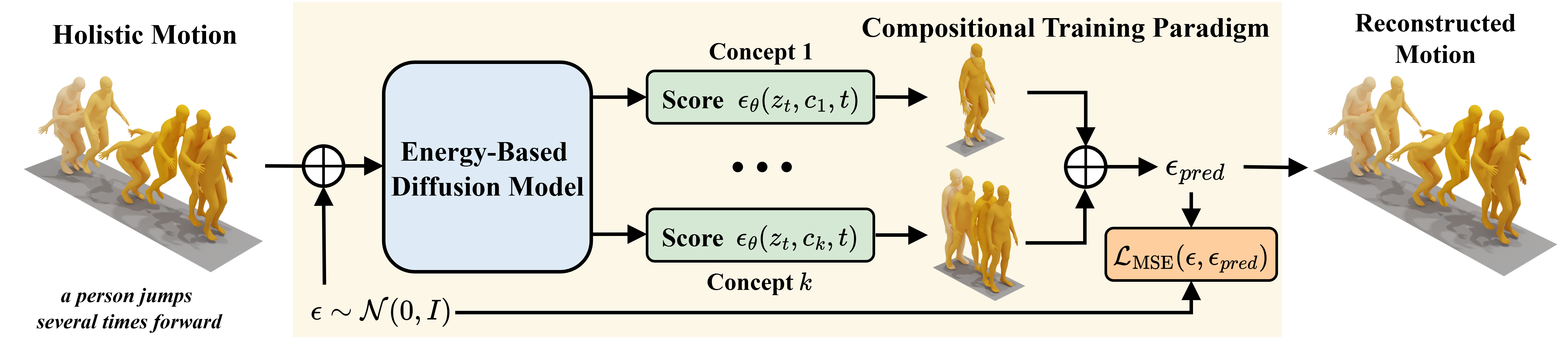}
\vspace{-7mm}
\caption{\textbf{Overview of our approach.} We propose \textsc{DeMoGen}, a compositional training paradigm that facilitates decompositional motion generation via an energy-based diffusion model. We learn to decompose the holistic motion into $K$ concepts. The energy functions of these concepts are aggregated to form the $\epsilon_{pred}$, which is subsequently trained to guide the denoising process. Energy aggregation can be achieved via a denoising network and a cross-attention, supporting either latent-aware or semantic-aware modeling~\ref{sec:demogen}. Furthermore, we investigate three variants (Section~\ref{sec:variants}), which specify distinct strategies for learning or utilizing $\{\vc_k\}_{k=1}^K$ within our approach.}
\label{fig:overview}
\vspace{-2mm}
\end{figure*}

In this section, we detail our approach, \textsc{DeMoGen}. We first introduce background knowledge on interpreting diffusion models and cross-attention modules as parameterizing energy functions in Section~\ref{sec:preliminary}. Next, we discuss how to use this interpretation to decompose a motion sequence into multiple composable motion concepts, which are then leveraged for compositional training in Section~\ref{sec:demogen}. We present three \textsc{DeMoGen} variants in Section~\ref{sec:variants}, reflecting different levels of compositional supervision. In Section~\ref{sec:decompml_dataset}, we introduce the DeCompML dataset. The overview of our approach is illustrated in Figure~\ref{fig:overview}.

\subsection{Energy Perspectives of Diffusion Models}
\label{sec:preliminary}
Denoising Diffusion Probabilistic Models~\cite{ho2020denoising} facilitate motion generation by progressively refining motion sequences corrupted with Gaussian noise. The model consists of a forward diffusion process and a reverse denoising process. The forward process is defined as a Markov chain that models the distribution $q(\vx_t|\vx_{t-1}) = \mathcal{N}(\vx_t;\sqrt{1-\beta_t}\vx_{t-1},\beta_tI)$, where the clean motion $\vx_0$ is gradually perturbed by Gaussian noise $\epsilon \sim \mathcal{N}(0,I)$ over $t$ steps. Then, a denoising network $\epsilon_\theta$ is trained to predict the noise at each time step $t$
\begin{equation}
\mathcal{L}_{\text{MSE}} =  \Vert \epsilon - \epsilon_\theta(\vx_t, t) \Vert^{2}.
\label{formula:diffusionloss}
\end{equation}
Conversely, the reverse process seeks to recover the original motion by iteratively removing the added noise. Specifically, given a sample $\vx_T$ at noise level $T$, the motion can be denoised through
\begin{equation}
\vx_{t-1} = \vx_t - \epsilon_\theta(\vx_t, t) + \mathcal{N}(0, \tilde{\beta}_tI),
\label{formula:denoising_process}
\end{equation}
where $\tilde{\beta}$ denotes the posterior variance. Following prior efforts~\cite{song2020score,du2019implicit,liu2022compositional,du2020compositional,welling2011bayesian}, the denoising network $\epsilon_\theta(\vx, t)$ can be interpreted as a score function $\nabla_\vx\log p(\vx)$, which, in turn, establishes a correspondence with the energy function, \ie, $\epsilon_\theta(\vx, t) \propto \nabla_\vx E_\theta(\vx)$, where $p(\vx) \sim e^{-E(\vx)}$. Thus, with the step size $\eta$, Equation~\ref{formula:denoising_process} is equivalent to
\begin{equation}
\vx_{t-1} = \vx_t - \eta \nabla_\vx E_\theta(\vx) + \mathcal{N}(0, \tilde{\beta}_tI).
\label{formula:diffusion_process}
\end{equation}
EnergyMoGen~\cite{zhang2025energymogen} defines the preceding expression as a latent-aware EBM and also interprets cross-attention in the denoising network as a semantic-aware EBM. This formulation draws inspiration from modern Hopfield networks~\cite{ramsauer2020hopfield,hoover2024energy,hu2024statistical,hu2024provably,park2024energy}, which model associative patterns and contextual relationships within text embeddings. These energy-based interpretations motivate us to employ energy-based diffusion models (both latent- and semantic-aware) to represent a holistic motion using a set of energy functions, where each function captures a specific motion concept, collectively enabling a decompositional understanding of motion.

\begin{figure}[t]
\begin{minipage}{0.47\textwidth}
\vspace{-2mm}
\begin{algorithm}[H]
    \centering
    \caption{Compositional Training Paradigm}
    \label{alg:training}
    \begin{algorithmic}[1]
        \STATE \textbf{Require}: \hspace{-0.1cm} Frozen VAE encoder $\mathcal{E}$, \hspace{-0.1cm} denoising network $\epsilon_{\theta}$, number of training data $N$, learning rate $\lambda$
        \FOR{$i = 1, \ldots, N$}
            \STATE $\vz^i \gets \mathcal{E}(\vx^i)$ \\
            \STATE {\color{gray}\small{// \textit{Initialize a Gaussian noise and a random time step}}} \\
            \STATE $\epsilon \sim \mathcal{N}(0, 1)$, $t \sim \mathcal{U}\{1,\ldots,T\}$ \\
            \STATE $\vz_t^i = \sqrt{{\bar{\alpha}}}_t\vz^i + \sqrt{1-{\bar{\alpha}}_t} \epsilon$ \hspace{0.45cm} {\color{gray}\small{//\ $\bar{\alpha}_t = \prod_{i=1}^{t} (1 - \beta_i)$}} \\
            \STATE {\color{gray}\small{// \textit{Compute compositional denoising scores}}}  \\
            \IF{ mod is \textbf{latent-aware} }
            \STATE $\epsilon_{pred} \gets \textstyle{\sum_{k=1}^{K}} \epsilon_\theta(\vz^i_t, \vc_k, t)$ \\ 
            \ENDIF \\
            \IF{ mod is \textbf{semantic-aware} }
            \STATE {\color{gray}\small{// \textit{Replace the cross-attention with DCA}}} \\
            \STATE $\epsilon_{pred} \gets \epsilon_\theta(\vz^i_t, \mC, t)$
            \ENDIF \\
            \STATE $\mathcal{L} = \text{MSE}(\epsilon, \epsilon_{pred})$ \hspace{1.6cm} {\color{gray}\small{// \textit{Optimization goal}}} \\
            \STATE $\theta \leftarrow \theta - \lambda \nabla_\theta \mathcal{L}$  \hspace{1.25cm} {\color{gray}\small{// \textit{Update the model weight}}} 
        \ENDFOR \\
    \end{algorithmic}
\end{algorithm}
\end{minipage}
\vspace{-4mm}
\end{figure}

\subsection{Decompositional Human Motion Generation with Energy-Based Diffusion Models}
\label{sec:demogen}
A well-performed text-to-motion model should not only synthesize text-aligned motions but also capture their underlying decompositional structure, \ie, how a motion is composed of multiple primitives. To achieve this, our \textsc{DeMoGen} reformulates the motion generation task from a decompositional perspective. We adopt a skeleton-aware latent diffusion architecture following~\cite{hong2025salad}. Specifically, given a motion $\vx \in \mathbb{R}^{L \times d_m}$, where $L$ denotes the motion length and $d_m$ is the feature dimensionality of each frame. We variationally encode the holistic motion $\vx$ into a latent representation $\vz$ using a motion encoder $\mathcal{E}$, and map it back to the motion space through a decoder $\mathcal{D}$.

Our key idea is to learn a set of energy functions to reconstruct a motion sequence using an energy-based diffusion model in a compositional manner. Driven by this insight and the close connection between diffusion models and EBMs discussed in Section~\ref{sec:preliminary}, we propose a compositional training paradigm.
Concretely, at noise level $t$, the noisy latent $\vz_t$ is associated with a set of text embeddings $\mC = \{\vc_k\}_{k=1}^K$, each corresponding to a target motion concept, where $K$ denotes the number of motion concepts decomposed from a holistic motion. Note that several strategies to obtain $\{\vc_k\}_{k=1}^K$ are detailed in Section~\ref{sec:variants}.

\begin{table*}[t]
    \centering
    \setlength{\tabcolsep}{5pt}
    \resizebox{0.96\linewidth}{!}{

    \begin{tabular}{l c c c c c c c}
    \toprule
    \multirow{2}{*}{Methods}  & \multicolumn{3}{c}{R-Precision $\uparrow$} & \multirow{2}{*}{FID $\downarrow$} & \multirow{2}{*}{MM-Dist $\downarrow$} & \multirow{2}{*}{Diversity $\rightarrow$} & \multirow{2}{*}{MModality $\uparrow$}\\

    \cline{2-4}
    ~ & Top-1 & Top-2 & Top-3 \\
    
    \midrule

        \textbf{Real motion}& \et{0.511}{.003} & \et{0.703}{.003} & \et{0.797}{.002} & \et{0.002}{.000} & \et{2.974}{.008} & \et{9.503}{.065} & -  \\
    \midrule
        MDM~\cite{tevet2022MDM} & \et{0.418}{.005} & \et{0.604}{.001} & \et{0.707}{.004} & \et{0.489}{.025} & \et{3.630}{.023} & \et{\underline{9.450}}{.066} & \etb{2.870}{1.11}  \\ 
    
        MLD~\cite{chen2022mld} & \et{0.481}{.003} & \et{0.673}{.003} & \et{0.772}{.002} & \et{0.473}{.013} & \et{3.196}{.010} & \et{9.724}{.082} & \et{2.413}{.079} \\

        ReMoDiffusion~\cite{zhang2023remodiffuse} & \et{0.510}{.005} & \et{0.698}{.006} & \et{0.795}{.004} & \et{0.103}{.004} & \et{2.974}{.016} & \et{9.018}{.075} & \et{1.795}{.043}  \\ 

        FineMoGen~\cite{zhang2023finemogen} & \et{0.504}{.002} & \et{0.690}{.002} & \et{0.784}{.004} & \et{0.151}{.008} & \et{2.998}{.008} & \et{9.263}{.094} & \et{\underline{2.696}}{.079}  \\ 

        EnergyMoGen~\cite{zhang2025energymogen} & \et{0.523}{.003} & \et{0.715}{.002} & \et{0.815}{.002} & \et{0.188}{.006} & \et{2.915}{.007} & \etb{9.488}{.099} & \et{2.205}{0.041} \\ 

        SALAD~\cite{hong2025salad} & \et{0.581}{.003} & \et{0.769}{.003} & \et{0.857}{.002} & \etb{0.076}{.002} & \et{2.649}{.009} & \et{9.696}{.096} & \et{1.751}{0.062} \\ 
        
        \midrule
        \rowcolor{gray!20}
        \noalign{\vspace{-0.8mm}}
        \multicolumn{8}{c}{\textit{\textbf{latent-aware}}} \\
        \textsc{DeMoGen-Exp} & \et{0.569}{.004} & \et{0.760}{.005} & \et{0.850}{.004} & \et{\underline{0.078}}{.003} & \et{2.708}{.012} & \et{9.774}{.099} & \et{1.944}{.054} \\
        
        \textsc{DeMoGen-SC} & \et{0.565}{.003} & \et{0.757}{.004} & \et{0.846}{.003} & \et{0.121}{.005} & \et{2.739}{.009} & \et{9.933}{.131} & \et{1.785}{.044} \\
        
        \textsc{DeMoGen-OSS} & \etb{0.588}{.004} & \etb{0.778}{.002} & \et{\underline{0.861}}{.003} & \et{0.092}{.003} & \et{\underline{2.625}}{.007} & \et{9.779}{.120} & \et{1.748}{.059}  \\ 
        
        \midrule
        \rowcolor{gray!20}
        \noalign{\vspace{-0.8mm}}
        \multicolumn{8}{c}{\textit{\textbf{semantic-aware}}} \\
        
        \textsc{DeMoGen-Exp} & \et{\underline{0.586}}{.005} & \et{\underline{0.776}}{.003} & \etb{0.863}{.002} & \et{0.116}{.008} & \etb{2.623}{.008} & \et{9.873}{057} & \et{1.785}{.046} \\ 
        
        \textsc{DeMoGen-SC} & \et{0.565}{.002} & \et{0.758}{.003} & \et{0.846}{.003} & \et{0.138}{.005} & \et{2.727}{.010} & \et{9.769}{.158} & \et{1.763}{.046} \\
        \textsc{DeMoGen-OSS} & \et{0.584}{.002} & \et{0.774}{.003} & \et{0.858}{.001} & \et{0.104}{.005} & \et{2.637}{.014} & \et{9.877}{.127} & \et{1.742}{.102} \\
        
    \bottomrule
    \end{tabular}
    }
    \vspace{-2mm}
    \caption{\textbf{Comparison with the state-of-the-art diffusion models on the test set of HumanML3D~\cite{guo2022generating}.} We quantitatively evaluate our approach across three variants under both latent-aware and semantic-aware settings. All metrics are obtained via a pretrained evaluation model from Guo~\etal~\cite{guo2022generating}. \textbf{Bold} and \underline{underlined} denote the best and second-best results, respectively.}
    \label{tab:HumanML}
    \vspace{-2mm}
\end{table*}

We further investigate the accommodation of our training paradigm to the two spectra of EBMs~\cite{zhang2025energymogen}, \ie, latent-aware and semantic-aware, under the same principle with minimal modifications:
\begin{itemize}
    \item\textit{Latent-aware EBM} is realized by the denoising network, which is trained to capture the composed energy distribution across these $K$ concepts using the objective:
    \begin{equation}
    \mathcal{L}_{\text{MSE}} =  \Vert \epsilon - \textstyle{\sum_{k=1}^{K}} \epsilon_\theta(\vz_t, \vc_k, t) \Vert^{2}.
    \label{formula:diffusionloss_latent}
    \end{equation}
    \item\textit{Semantic-aware EBM} parameterizes the energy function via the cross-attention module. We propose a Decompositional Cross-Attention (DCA) mechanism that divides the attention computation into parallel branches. It performs $K$ distinct attention operations with different key–value pairs and aggregates their outputs, formulated as
    \begin{equation}
    \text{DCA}(\vz_t, \mC) = \textstyle{\sum_{k=1}^{K}} \text{CA}(\vz_t, \vc_k),
    \label{formula:dca}
    \end{equation}
    where CA indicates the standard cross-attention. We then optimize the entire model using the general optimization goal $\mathcal{L}_{\text{MSE}} =  \Vert \epsilon - \epsilon_\theta(\vz_t, \mC, t) \Vert^{2}$.
\end{itemize}
With these energy functions effectively reconstructed, we can directly extract the specific and desired motion concepts from the learned decompositional space and further operate over them, thereby achieving motion decomposition and concept recombination. Furthermore, this compositional training paradigm can also improve the text-to-motion generation. We provide the pseudocode for our approach in Algorithm~\ref{alg:training}. The pseudocodes for sampling are provided in Appendix~\ref{sec:supp_implementation}.

\subsection{\textbf{\textsc{DeMoGen}} Variants}
\label{sec:variants}
To comprehensively study the proposed training paradigm, we present three variants that reflect different levels of compositional supervision. For convenience, we denote $\mC^P = \{\vc^P_k\}_{k=1}^K$ as the set of predefined decomposed text embeddings from our DeCompML dataset (Section~\ref{sec:decompml_dataset}). We present extensive ablation studies of these three variants in the appendix.

\vspace{0.5mm}
\noindent\textbf{\textsc{DeMoGen-Exp}} provides \textbf{\textit{explicit supervision}} by indicating the correspondence between the holistic motion and its decomposed textual descriptions, where $\mC = \mC^P$. Relying solely on decomposed textual descriptions for training may lead to inaccurate text–motion mapping and limit the model’s performance. To address this issue, we propose a text mixing strategy, that is, in each training batch, $\tau \times 100\%$ of the motion samples are trained with the original textual descriptions instead of the decomposed ones.

\vspace{0.5mm}
\noindent\textbf{\textsc{DeMoGen-OSS}} is trained with \textit{\textbf{orthogonal self-supervision}}. Specifically, given an original text embedding $\vc \in \mathbb{R}^{L_c \times d_c}$, we partition it along the dimensional axis into $K$ sub-embeddings $\hat{\mC} = \{\hat{\vc}_k\}_{k=1}^K$, each of size $d_c / K$. We feed $\hat{\mC}$ into a fully connected layer to produce $\mC$. We promote orthogonality among the components in $\mC$ using an orthogonalization loss $\mathcal{L}_\text{Ortho}$, following~\cite{su2024compositional}. Training proceeds in a fully self-supervised manner, without relying on any predefined text embeddings. The training objective can be written as
$\mathcal{L} = \mathcal{L}_\text{MSE} + \alpha_{o} \mathcal{L}_\text{Ortho}$,
where $\alpha_{o}$ is the weight of the orthogonalization loss.

\vspace{0.5mm}
\noindent\textbf{\textsc{DeMoGen-SC}} splits the original text embedding following the same procedure as \textsc{DeMoGen-OSS}, but uses two-layer transformers to obtain $\mC$. We employ \textbf{\textit{semantic consistency supervision}} by encouraging the consistency property between the partitioned and decomposed text embeddings, which is  $\mathcal{L}_{SC} = \mathcal{L}_1^\text{smooth}(\mC^P, \mC)$. Furthermore, we also use $\mathcal{L}_\text{Ortho}$ to encourage disentanglement. Thus, the overall optimization goal is $\mathcal{L} = \mathcal{L}_\text{MSE} + \alpha_{sc} \mathcal{L}_\text{SC}  + \alpha_{o} \mathcal{L}_\text{Ortho}$, with an additional weight hyper-parameter $\alpha_{sc}$.

During inference, \textsc{DeMoGen-OSS} and \textsc{DeMoGen-SC} operate without the need for predefined decompositional texts, thus enabling unguided motion concept discovery (Section~\ref{sec:exp_decomp_comp}). We further observe that \textsc{DeMoGen-OSS} benefits the diversity for generating holistic motions (Section~\ref{sec:exp_discuss}). In contrast, \textsc{DeMoGen-Exp} leverages decompositional texts to achieve more accurate and diverse motion decomposition, while also exhibiting superior compositional generation performance (Section~\ref{sec:exp_decomp_comp}). It's worth noting that \textsc{DeMoGen-Exp} supports single textual input for the text-to-motion generation task by duplicating the text to satisfy the required decompositional conditioning.

\begin{figure*}[tp]
\centering
\includegraphics[width=0.99\textwidth]{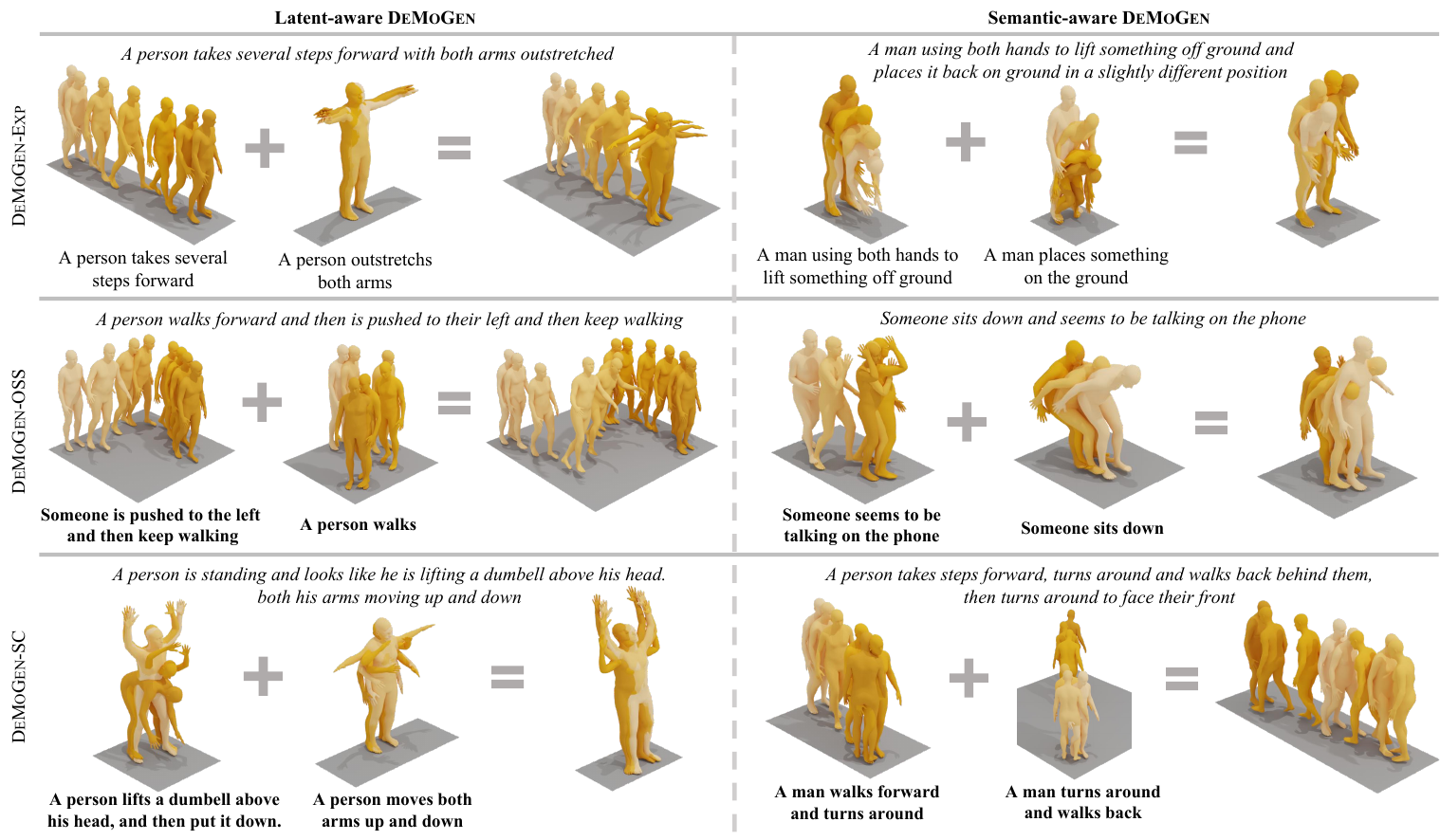}
\vspace{-3mm}
\caption{\textbf{Text-to-motion generation with decompositional understanding.} Given a complete textual description (in italics above the result), our approach first infers the motion concepts and further composes them to synthesize the holistic motion that matches the text. Notably, \textsc{DeMoGen-OSS} and \textsc{DeMoGen-SC} discover motion concepts without the aid of decomposed text. However, for clarity, we manually annotate each concept in \textbf{bold}. More visual results can be found on the \href{https://jiro-zhang.github.io/DeMoGen/}{project page}.}
\vspace{-2mm}
\label{fig:exp_text_to_motion}
\end{figure*}

\subsection{DeCompML Dataset}
\label{sec:decompml_dataset}
To support explicit and semantic consistency supervisions in \textsc{DeMoGen-Exp} and \textsc{DeMoGen-SC} mentioned above, we relabel the HumanML3D~\cite{guo2022generating} dataset. HumanML3D collects 14,616 motion sequences, each paired with 3–4 textual descriptions summarizing the entire motion. We relabel these texts by splitting each into two sentences, which respectively depict different decompositional motion concepts within the holistic sequence. For example, a description such as ``\textit{someone runs forward, turns around, and then walks backward} '' is extended into ``\textit{someone runs forward and turns around} '' and ``\textit{someone walks backward} ''. This process is fully automated by prompting a large language model (\ie, GPT-4.1~\cite{openai2024gpt4}) with carefully crafted instructions, constraints, \etc.
Considering that HumanML3D~\cite{guo2022generating} already applies data augmentation strategies such as mirroring, our DeCompML dataset ultimately comprises 87,384 decomposed text groups (each containing two sentences, 174,768 in total). This dataset can serve as a new benchmark for compositional motion generation (Section~\ref{sec:exp_decomp_comp}).

With the relabeled textual descriptions, we sample decomposed motions in the HumanML3D training set via the learned \textsc{DeMoGen-Exp}, and select 15,000 high-quality motion sequences as extended motion data. 
We demonstrate that jointly training on both the HumanML3D and the 15,000 decomposed motion–text pairs from DeCompML can improve text-to-motion generation (Section~\ref{sec:exp_discuss}). 

We provide the designed prompt, statistics of DeCompML, and ablative experiments on the number of textual splits in the appendix.
\begin{figure*}[tp]
\centering
\includegraphics[width=0.98\textwidth]{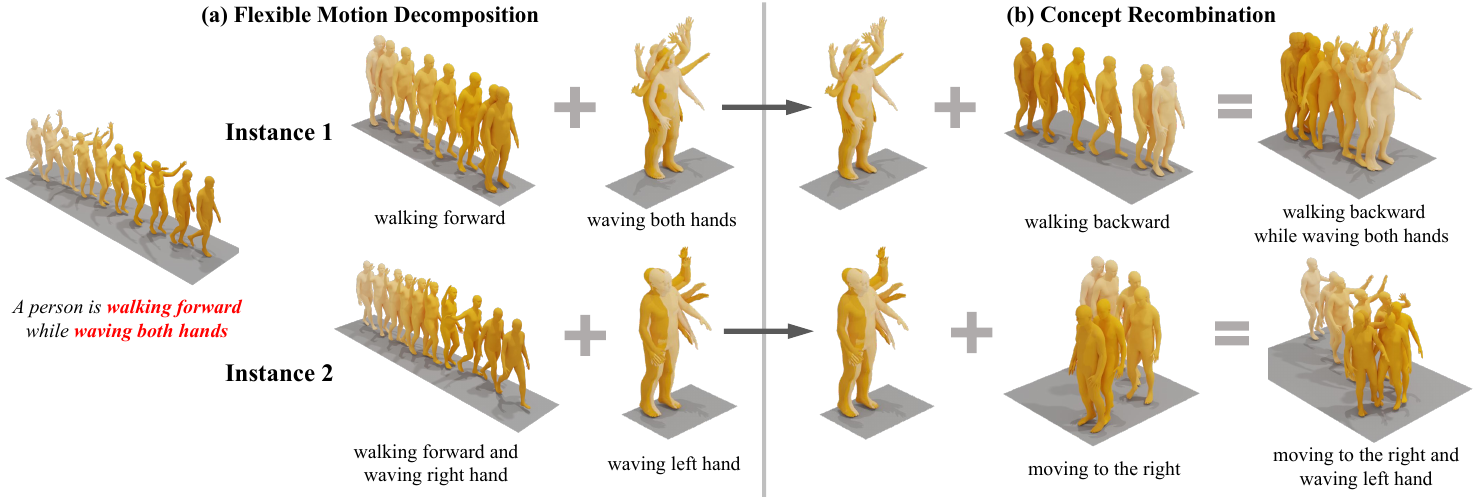}
\vspace{-3mm}
\caption{\textbf{Motion decomposition and recombination.} We demonstrate that our method can infer diverse motion concepts from a complex motion sequence, conditioned on different decompositional text prompts. Our approach also exhibits the ability to recombine the inferred concepts with others to generate novel motions.}
\vspace{-2mm}
\label{fig:exp_decompositional}
\end{figure*} 

\section{Experiments}
\label{sec:experiments}
In this section, we evaluate our approach across four tasks: text-to-motion generation (Section~\ref{sec:exp_text2motion}), decompositional and compositional motion generation (Section~\ref{sec:exp_decomp_comp}), as well as multi-concept motion generation (Section~\ref{sec:exp_decomp_comp}). Analysis and discussion are provided in Section~\ref{sec:exp_discuss}. More detailed information on datasets and evaluation metrics can be found in the appendix.

\subsection{Datasets and Evaluation Metrics}
\label{sec:exp_data}
\vspace{1mm}
\noindent\textbf{Datasets.}
We conduct experiments on several benchmark datasets across different tasks. HumanML3D~\cite{guo2022generating} and our DeCompML collectively support text-to-motion generation and decompositional motion generation by providing both holistic and decomposed motion–text pairs. To examine our approach under complex scenarios involving multiple motion concepts, we adopt the MTT~\cite{petrovich24stmc} dataset for multi-concept generation and motion composition. Notably, the decomposed annotations of DeCompML also make it a suitable benchmark for motion composition tasks.

\begin{table}
\centering
\setlength{\tabcolsep}{3pt}
\resizebox{1.07\linewidth}{!}{
\begin{tabular}{l|cc|cc|cc}
\toprule
\multirow{2}{*}{Methods} & \multicolumn{2}{c|}{R-Presicion} & \multicolumn{2}{c|}{TMR-Score $\uparrow$} & \multirow{2}{*}{\text{FID} $\downarrow$} & Transition \\
& R@1 $\uparrow$ & R@3 $\uparrow$ & M2T & M2M & & distance $\downarrow$ \\

\midrule
\multicolumn{7}{c}{Multi-concept motion generation (single text)} \\
\midrule
MotionDiffuse~\cite{zhang2022motiondiffuse}& 10.9 & 21.3 & 0.558 & 0.546 & 0.621 & 1.9 \\   
MDM~\cite{tevet2022MDM} & 9.5  & 19.7 & 0.556 & 0.549 & 0.666 & 2.5 \\
ReModiffuse~\cite{zhang2023remodiffuse} & 7.4 & 18.3 & 0.531 & 0.534 & 0.699 & 3.3 \\
FineMoGen~\cite{zhang2023finemogen} & 5.4 & 11.7 & 0.504 & 0.533 & 0.948 & 9.4 \\
MLD~\cite{chen2022mld} & 10.5 & 22.3 & 0.559 & 0.552 & 0.685 & 2.4 \\
EMG~\cite{zhang2025energymogen} & 12.7  & 25.4 & 0.570 & 0.562 & 0.592 & 2.7 \\
EMG+AGD~\cite{zhang2025energymogen} & 14.0  & 26.3 & 0.570 & 0.560 & 0.587 & 2.7 \\
\midrule
\textsc{DeMoGen-SC}$^\dag$ & 14.3 & 29.7 & 0.578 & 0.568 & 0.585 & 2.7 \\
\textsc{DeMoGen-OSS}$^\dag$ & 14.9 & 29.5 & 0.584 & 0.574 & 0.580 & 2.6 \\
\midrule
\multicolumn{7}{c}{Compositional motion generation (multiple texts)} \\
\midrule
EMG+SEF~\cite{zhang2025energymogen} & 15.9  & 28.0 & 0.591 & 0.567 & 0.604 & 1.6 \\ \midrule
\textsc{DeMoGen-Exp}$^\dag$ & 16.2  & 31.9 & 0.597 & 0.570 & 0.621 & 1.6 \\
\bottomrule        
\end{tabular}
}
\vspace{-2mm}
\caption{\textbf{Quantitative comparison on MTT~\cite{petrovich24stmc}}. The metrics are computed following STMC~\cite{petrovich24stmc} and EnergyMoGen (EMG)~\cite{zhang2025energymogen}. $\dag$ indicates the latent-aware setting. The results of the semantic-aware model are provided in the appendix.}
\vspace{-4mm}
\label{tab:composition}
\end{table}

\vspace{1mm}
\noindent\textbf{Evaluation Metrics.}
To evaluate the performance of our approach, we report R-Precision, MM-Dist, FID, Diversity, and MModality for text-to-motion generation following Guo~\etal~\cite{guo2022generating}. We employ the same metrics as text-to-motion for motion composition on DeCompML. On MTT~\cite{petrovich24stmc}, R-Precision, TMR-Score, FID, and transition distance are used as quantitative metrics for both compositional and multi-concept generation.

\subsection{Implementation Details}
We use the VAE initialized with pretrained weights from~\cite{hong2025salad}.
We train the latent-aware and semantic-aware \textsc{DeMoGen} using the AdamW optimizer with a batch size of 64 for 500 epochs. The initial learning rate is set to 2e-4 and decayed to 2e-5 after 50K iterations. We set $K$=2 to model each holistic motion with two motion concepts, and following~\cite{su2024compositional}, we compute the final energy distribution by averaging $K$ energy scores. For all three variants of \textsc{DeMoGen}, the text replacement rate $\tau$ is set to 0.7$\times 100\%$ and $\alpha_{sc}$ is set to 1.0. The orthogonal loss weight $\alpha_{o}$ is configured as 2.0 and 1.0 for latent-aware and semantic-aware training, respectively. 
For compositional and multi-concept motion generation, we directly use the model pre-trained
on HumanML3D~\cite{guo2022generating} and evaluate it on MTT~\cite{petrovich24stmc} and our DeCompML.
Ablation studies on these hyperparameters, along with additional implementation details, are provided in the appendix.

\subsection{Text-to-Motion Generation}
\label{sec:exp_text2motion}
\noindent\textbf{Quantitative results.}
We compare our approach with recent state-of-the-art (SOTA) methods, \eg, SALAD~\cite{hong2025salad} and EnergyMoGen~\cite{zhang2025energymogen}, on the HumanML3D~\cite{guo2022generating} test set. The results are summarized in Table~\ref{tab:HumanML}, from which we find that: \textit{i)} By promoting the decompositional modeling, our compositional training paradigm effectively improves text-to-motion generation.
This is particularly shown by \textsc{DeMoGen-OSS}'s superior performance in terms of text-motion consistency, achieved while maintaining comparable FID scores against baselines. \textit{ii)} The latent-aware \textsc{DeMoGen} holds an advantage in generating smoother motion compared to the semantic-aware models, as reflected by its lower FID score. Note that our approach can be seamlessly integrated with other diffusion methods, \eg, MLD~\cite{chen2022mld} and MotionDiffuse~\cite{zhang2022motiondiffuse}, and results are provided in Appendix~\ref{sec:supp_other_models}.

\vspace{1mm}
\noindent\textbf{Motion Generation with Decompositional Understanding.}
In Figure~\ref{fig:exp_text_to_motion}, we show how our approach infers a set of motion concepts from the given textual descriptions. These concepts can subsequently be composed to generate the target motion sequence. \textsc{DeMoGen-Exp} requires the decompositional text from DeCompML as explicit input, while \textsc{DeMoGen-OSS} and \textsc{DeMoGen-SC} enable the motion generation without such decompositional guidance. As expected, all three variants effectively discover the compositional motion concepts intrinsic to the text-motion correspondence. These qualitative observations reinforce the quantitative results, further showing that our training paradigm endows the models with the capability of decompositional understanding.

\begin{table}[t]
    \centering\setlength{\tabcolsep}{5pt}
    \resizebox{0.48\textwidth}{!}{
    \begin{tabular}{l|ccc}
    \toprule
        Methods & Top-1 $\uparrow$ & FID $\downarrow$  & MM-Dist $\downarrow$ \\
        \midrule
        EnergyMoGen & \et{0.326}{.009} & \et{2.502}{.028} & \et{4.348}{.008} \\
        SALAD & \et{0.342}{.007} & \et{1.267}{.024} & \et{4.282}{.018} \\ \midrule 
        Ours (latent) & \et{0.559}{.004} & \etb{0.089}{.005} & \et{2.758}{.014} \\
        Ours (semantic) & \etb{0.567}{.004} & \et{0.102}{.007} & \etb{2.719}{.009} \\
        \bottomrule
    \end{tabular}
    }
    \vspace{-2mm}
    \caption{\textbf{Motion composition on DecompML.} We use the pretrained text-to-motion model on HumanML3D for quantitative evaluation. As a baseline, SALAD~\cite{hong2025salad} is modified to incorporate the compositional strategy from EnergyMoGen~\cite{zhang2025energymogen}.}
    \vspace{-2mm}
    \label{tab:exp_decompml}
\end{table}

\begin{figure}[tp]
\centering
\includegraphics[width=0.48\textwidth]{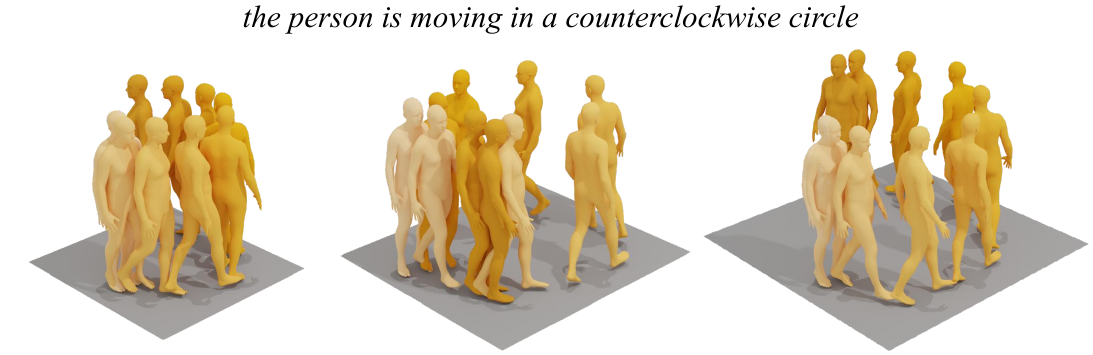}
\vspace{-7mm}
\caption{\textbf{Motion Diversity.} From left to right, we visualize the motions generated by \textsc{DeMoGen-OSS} under a latent-aware setting for inferred concept 1, concept 2, and their combination.}
\vspace{-4mm}
\label{fig:exp_diversity}
\end{figure} 

\subsection{Motion Decomposition and Composition}
\label{sec:exp_decomp_comp}
\noindent\textbf{Decomposition and Recombination.}
In Figure~\ref{fig:exp_decompositional}(a), we illustrate how our approach (latent-aware \textsc{DeMoGen-Exp}), decomposes a complex motion into diverse and semantically interpretable sets of motion concepts guided by different decompositional texts. These results highlight the flexibility of our model in performing motion decomposition. Moreover, by recombining the decomposed concepts with others, our method is able to synthesize novel motions (as shown in Figure~\ref{fig:exp_decompositional}(b)) that do not appear in the training data. Collectively, these findings validate the effectiveness of our compositional training paradigm. More visual results of \textsc{DeMoGen-OSS} and \textsc{DeMoGen-SC} can be found on the \href{https://jiro-zhang.github.io/DeMoGen/}{project page}.

\vspace{1mm}
\noindent\textbf{Composition and Multi-concept generation.}
Table~\ref{tab:composition} presents the results of our approach against SOTA diffusion models on MTT. Since \textsc{DeMoGen-Exp} is trained directly with multiple textual descriptions, we use it to evaluate compositional motion generation. In contrast, the other two variants are more suitable for multi-concept generation. Our approach significantly outperforms prior SOTA models on both tasks.
The results in Table~\ref{tab:composition} correspond to the latent-aware setting of \textsc{DeMoGen}, which achieves better performance than EnergyMoGen across key evaluation metrics, such as R-Precision, TMR-Score, and FID.

Furthermore, our DeCompML dataset enables benchmarking for motion composition. Specifically, we construct compositional motion pairs by combining the textual annotations from DeCompML with motion samples from the HumanML3D~\cite{guo2022generating} test set. We then compare the generated results against the corresponding HumanML3D ground-truth motion using the same evaluation model in text-to-motion. As shown in Table~\ref{tab:exp_decompml}, our \textsc{DeMoGen-Exp} exhibits substantial performance advantages over prior approaches. Additional experimental results and ablation studies are provided in the appendix.

\begin{table}[t]
    \centering\setlength{\tabcolsep}{5pt}
    \resizebox{0.48\textwidth}{!}{
    \begin{tabular}{l|ccc}
    \toprule
        Methods & Top-1 $\uparrow$ & FID $\downarrow$  & MM-Dist $\downarrow$ \\
        \midrule
        EnergyMoGen & \et{0.523}{.003} & \et{0.188}{.006} & \et{2.915}{.007} \\
        EnergyMoGen$^*$ & \et{0.526}{.004} & \et{0.147}{.004} & \et{2.884}{.009} \\ \midrule 
        SALAD & \et{0.581}{.003} & \et{0.076}{.002} & \et{2.649}{.009} \\ 
        SALAD$^*$ & \et{0.580}{.004} & \et{0.060}{.005} & \et{2.651}{.014} \\
        \bottomrule
    \end{tabular}
    }
    \vspace{-2mm}
    \caption{\textbf{Text-to-motion evaluation on extended HumanML3D.} $^*$ indicates the finetuned model. Additional metrics and results of other methods are provided in Appendix~\ref{sec:supp_t2m_decompml}.}
    \vspace{-2mm}
    \label{tab:exp_extend}
\end{table}

\subsection{Discussion}
\label{sec:exp_discuss}

\noindent\textbf{Motion Diversity.} We observe that many motions in HumanML3D are simple and not amenable to further decomposition. For such motions, both \textsc{DeMoGen-OSS} and \textsc{DeMoGen-SC} can generate different variations in their expression. As illustrated in Figure~\ref{fig:exp_diversity}, the synthesized motions exhibit diversity in terms of circle size and displacement. These results validate the effectiveness of two loss terms, \ie, $\mathcal{L}_\text{SC}$ and $\mathcal{L}_\text{Ortho}$. Ablation studies on them are provided in Appendix~\ref{sec:supp_demogen_oss} and~\ref{sec:supp_demogen_sc}.

\vspace{1mm}
\noindent\textbf{Experiments on DeCompML.} 
We finetune the pretrained text-to-motion models by incorporating data from both HumanML3D and DeCompML. Table~\ref{tab:exp_extend} shows that finetuning SALAD significantly reduces FID by around 21\%, while Top1 and MM-Dist remain largely unchanged. Integrating DeCompML improves all metrics for EnergyMoGen.
The findings indicate that our compositional training paradigm facilitates the generation of high-quality decomposed motions, which can be effectively used to augment the dataset.

\section{Conclusion}
In this paper, we propose a compositional training paradigm, \textsc{DeMoGen}, and explore three regimes of compositional supervision for decompositional motion generation. Our approach is able to effectively decompose human motion into distinct motion concepts, which can be flexibly recombined with external concepts to synthesize novel motions. We introduce the DeCompML dataset to facilitate compositional training, which can also serve as a benchmark for motion composition. Experimental results demonstrate the superior performance of our approach across various tasks, particularly in compositional and multi-concept motion generation. Moreover, our investigation reveals that leveraging the decomposed motions as additional training data still yields benefits for text-to-motion performance.

{
    \small
    \bibliographystyle{ieeenat_fullname}
    \bibliography{main}

@string{nips={Advances in Neural Information Processing Systems (NeurIPS)}}

@string{icml={International Conference on Machine Learning (ICML)}}

@string{iclr={International Conference on Learning Representations (ICLR)}}

@string{cvpr={Proceedings of the Conference on Computer Vision and Pattern Recognition (CVPR)}}

@string{cvprw={Proceedings of the Conference on Computer Vision and Pattern Recognition Workshops (CVPRW)}}

@string{iccv={Proceedings of the International Conference on Computer Vision (ICCV)}}

@string{eccv={Proceedings of the European Conference on Computer Vision (ECCV)}}

@string{aaai={Proceedings of the AAAI Conference on Artificial Intelligence}}

@string{tdv={International Conference on 3D Vision (3DV)}}

@string{acmmm={Proceedings of the ACM International Conference on Multimedia (ACMMM)}}

@InProceedings{Zhang_2023_CVPR,
    author    = {Zhang, Jianrong and Zhang, Yangsong and Cun, Xiaodong and Zhang, Yong and Zhao, Hongwei and Lu, Hongtao and Shen, Xi and Shan, Ying},
    title     = {Generating Human Motion From Textual Descriptions With Discrete Representations},
    booktitle = cvpr,
    year      = {2023},
}

@inproceedings{zhang2025energymogen,
  title={Energymogen: Compositional human motion generation with energy-based diffusion model in latent space},
  author={Zhang, Jianrong and Fan, Hehe and Yang, Yi},
  booktitle=cvpr,
  year={2025}
}

@inproceedings{guo2024momask,
  title={Momask: Generative masked modeling of 3d human motions},
  author={Guo, Chuan and Mu, Yuxuan and Javed, Muhammad Gohar and Wang, Sen and Cheng, Li},
  booktitle=cvpr,
  year={2024}
}

@inproceedings{pinyoanuntapong2024mmm,
  title={MMM: Generative Masked Motion Model}, 
  author={Ekkasit Pinyoanuntapong and Pu Wang and Minwoo Lee and Chen Chen},
  booktitle=cvpr,
  year={2024},
}

@inproceedings{petrovich2023tmr,
  title={TMR: Text-to-motion retrieval using contrastive 3D human motion synthesis},
  author={Petrovich, Mathis and Black, Michael J and Varol, G{\"u}l},
  booktitle=iccv,
  year={2023}
}

@article{song2020score,
  title={Score-based generative modeling through stochastic differential equations},
  author={Song, Yang and Sohl-Dickstein, Jascha and Kingma, Diederik P and Kumar, Abhishek and Ermon, Stefano and Poole, Ben},
  journal={arXiv preprint arXiv:2011.13456},
  year={2020}
}

@article{zhang2023finemogen,
  title={FineMoGen: Fine-Grained Spatio-Temporal Motion Generation and Editing},
  author={Zhang, Mingyuan and Li, Huirong and Cai, Zhongang and Ren, Jiawei and Yang, Lei and Liu, Ziwei},
  journal=nips,
  year={2023}
}

@inproceedings{karunratanakul2023gmd,
  title={Guided Motion Diffusion for Controllable Human Motion Synthesis},
  author={Karunratanakul, Korrawe and Preechakul, Konpat and Suwajanakorn, Supasorn and Tang, Siyu},
  booktitle=iccv,
  year={2023}
}

@InProceedings{dabral2022mofusion,
      title={MoFusion: A Framework for Denoising-Diffusion-based Motion Synthesis},
      author={Rishabh Dabral and Muhammad Hamza Mughal and Vladislav Golyanik and Christian Theobalt},
      booktitle=cvpr,
      year={2023}
}

@article{zhang2023remodiffuse,
  title={ReMoDiffuse: Retrieval-Augmented Motion Diffusion Model},
  author={Zhang, Mingyuan and Guo, Xinying and Pan, Liang and Cai, Zhongang and Hong, Fangzhou and Li, Huirong and Yang, Lei and Liu, Ziwei},
  journal={arXiv},
  year={2023}
}

@article{jiang2023motiongpt,
  title={MotionGPT: Human Motion as a Foreign Language},
  author={Jiang, Biao and Chen, Xin and Liu, Wen and Yu, Jingyi and Yu, Gang and Chen, Tao},
  journal={arXiv},
  year={2023}
}

@InProceedings{chang2022maskgit,
  title = {MaskGIT: Masked Generative Image Transformer},
  author={Huiwen Chang and Han Zhang and Lu Jiang and Ce Liu and William T. Freeman},
  booktitle = cvpr,
  year = {2022}
}

@InProceedings{brown2020gpt3,
  title={Language models are few-shot learners},
  author={Brown, Tom and Mann, Benjamin and Ryder, Nick and Subbiah, Melanie and Kaplan, Jared D and Dhariwal, Prafulla and Neelakantan, Arvind and Shyam, Pranav and Sastry, Girish and Askell, Amanda and others},
  booktitle = nips,
  year={2020}
}

@article{liu2021learning,
  title={Learning to compose visual relations},
  author={Liu, Nan and Li, Shuang and Du, Yilun and Tenenbaum, Josh and Torralba, Antonio},
  journal=nips,
  year={2021}
}

@article{hoover2024energy,
  title={Energy transformer},
  author={Hoover, Benjamin and Liang, Yuchen and Pham, Bao and Panda, Rameswar and Strobelt, Hendrik and Chau, Duen Horng and Zaki, Mohammed and Krotov, Dmitry},
  journal=nips,
  year={2024}
}

@article{ramsauer2020hopfield,
  title={Hopfield networks is all you need},
  author={Ramsauer, Hubert and Sch{\"a}fl, Bernhard and Lehner, Johannes and Seidl, Philipp and Widrich, Michael and Adler, Thomas and Gruber, Lukas and Holzleitner, Markus and Pavlovi{\'c}, Milena and Sandve, Geir Kjetil and others},
  journal={arXiv preprint arXiv:2008.02217},
  year={2020}
}

@article{du2021unsupervised,
  title={Unsupervised learning of compositional energy concepts},
  author={Du, Yilun and Li, Shuang and Sharma, Yash and Tenenbaum, Josh and Mordatch, Igor},
  journal=nips,
  year={2021}
}

@article{du2019implicit,
  title={Implicit generation and modeling with energy based models},
  author={Du, Yilun and Mordatch, Igor},
  journal=nips,
  year={2019}
}

@article{lecun2006tutorial,
  title={A tutorial on energy-based learning},
  author={LeCun, Yann and Chopra, Sumit and Hadsell, Raia and Ranzato, M and Huang, Fujie and others},
  journal={Predicting structured data},
  year={2006}
}

@inproceedings{
zhao2017energybased,
title={Energy-based Generative Adversarial Networks},
author={Junbo Zhao and Michael Mathieu and Yann LeCun},
booktitle=iclr,
year={2017}
}

@inproceedings{sohl2015deep,
  title={Deep unsupervised learning using nonequilibrium thermodynamics},
  author={Sohl-Dickstein, Jascha and Weiss, Eric and Maheswaranathan, Niru and Ganguli, Surya},
  booktitle=icml,
  year={2015}
}

@inproceedings{shen2020interpreting,
  title={Interpreting the latent space of gans for semantic face editing},
  author={Shen, Yujun and Gu, Jinjin and Tang, Xiaoou and Zhou, Bolei},
  booktitle=cvpr,
  year={2020}
}

@inproceedings{peebles2020hessian,
  title={The hessian penalty: A weak prior for unsupervised disentanglement},
  author={Peebles, William and Peebles, John and Zhu, Jun-Yan and Efros, Alexei and Torralba, Antonio},
  booktitle=eccv,
  year={2020}
}

@inproceedings{singh2019finegan,
  title={Finegan: Unsupervised hierarchical disentanglement for fine-grained object generation and discovery},
  author={Singh, Krishna Kumar and Ojha, Utkarsh and Lee, Yong Jae},
  booktitle=cvpr,
  year={2019}
}

@inproceedings{preechakul2022diffusion,
  title={Diffusion autoencoders: Toward a meaningful and decodable representation},
  author={Preechakul, Konpat and Chatthee, Nattanat and Wizadwongsa, Suttisak and Suwajanakorn, Supasorn},
  booktitle=cvpr,
  year={2022}
}

@article{harkonen2020ganspace,
  title={Ganspace: Discovering interpretable gan controls},
  author={H{\"a}rk{\"o}nen, Erik and Hertzmann, Aaron and Lehtinen, Jaakko and Paris, Sylvain},
  journal=nips,
  year={2020}
}

@article{hu2024statistical,
  title={On statistical rates and provably efficient criteria of latent diffusion transformers (dits)},
  author={Hu, Jerry Yao-Chieh and Wu, Weimin and Li, Zhuoru and Pi, Sophia and Song, Zhao and Liu, Han},
  journal=nips,
  year={2024}
}

@article{hu2024provably,
  title={Provably optimal memory capacity for modern hopfield models: Transformer-compatible dense associative memories as spherical codes},
  author={Hu, Jerry Yao-Chieh and Wu, Dennis and Liu, Han},
  journal={arXiv preprint arXiv:2410.23126},
  year={2024}
}

@article{du2024compositional,
  title={Compositional generative modeling: A single model is not all you need},
  author={Du, Yilun and Kaelbling, Leslie},
  journal={arXiv preprint arXiv:2402.01103},
  year={2024}
}

@article{park2024energy,
  title={Energy-based cross attention for bayesian context update in text-to-image diffusion models},
  author={Park, Geon Yeong and Kim, Jeongsol and Kim, Beomsu and Lee, Sang Wan and Ye, Jong Chul},
  journal=nips,
  year={2024}
}

@inproceedings{hong2025salad,
  title={SALAD: Skeleton-aware Latent Diffusion for Text-driven Motion Generation and Editing},
  author={Hong, Seokhyeon and Kim, Chaelin and Yoon, Serin and Nam, Junghyun and Cha, Sihun and Noh, Junyong},
  booktitle=cvpr,
  year={2025}
}

@inproceedings{welling2011bayesian,
  title={Bayesian learning via stochastic gradient Langevin dynamics},
  author={Welling, Max and Teh, Yee W},
  booktitle=icml,
  year={2011}
}

@inproceedings{du2020compositional,
  title={Compositional visual generation with energy based models},
  author={Du, Yilun and Li, Shuang and Mordatch, Igor},
  booktitle=nips,
  year={2020}
}

@article{openai2024gpt4,
  title={GPT-4 Technical Report},
  author={OpenAI},
  journal={arXiv preprint arXiv:2303.08774},
  year={2024}
}

@article{su2024compositional,
  title={Compositional image decomposition with diffusion models},
  author={Su, Jocelin and Liu, Nan and Wang, Yanbo and Tenenbaum, Joshua B and Du, Yilun},
  journal={arXiv preprint arXiv:2406.19298},
  year={2024}
}

@article{pinyoanuntapong2024controlmm,
  title={ControlMM: Controllable Masked Motion Generation},
  author={Pinyoanuntapong, Ekkasit and Saleem, Muhammad Usama and Karunratanakul, Korrawe and Wang, Pu and Xue, Hongfei and Chen, Chen and Guo, Chuan and Cao, Junli and Ren, Jian and Tulyakov, Sergey},
  journal={arXiv preprint arXiv:2410.10780},
  year={2024}
}

@inproceedings{liu2022compositional,
  title={Compositional visual generation with composable diffusion models},
  author={Liu, Nan and Li, Shuang and Du, Yilun and Torralba, Antonio and Tenenbaum, Joshua B},
  booktitle=eccv,
  year={2022}
}

@inproceedings{liu2023unsupervised,
  title={Unsupervised compositional concepts discovery with text-to-image generative models},
  author={Liu, Nan and Du, Yilun and Li, Shuang and Tenenbaum, Joshua B and Torralba, Antonio},
  booktitle=iccv,
  year={2023}
}

@article{garipov2023compositional,
  title={Compositional sculpting of iterative generative processes},
  author={Garipov, Timur and De Peuter, Sebastiaan and Yang, Ge and Garg, Vikas and Kaski, Samuel and Jaakkola, Tommi},
  journal=nips,
  year={2023}
}

@inproceedings{shi2023exploring,
  title={Exploring compositional visual generation with latent classifier guidance},
  author={Shi, Changhao and Ni, Haomiao and Li, Kai and Han, Shaobo and Liang, Mingfu and Min, Martin Renqiang},
  booktitle=cvpr,
  year={2023}
}

@inproceedings{li2022stylet2i,
  title={Stylet2i: Toward compositional and high-fidelity text-to-image synthesis},
  author={Li, Zhiheng and Min, Martin Renqiang and Li, Kai and Xu, Chenliang},
  booktitle=cvpr,
  year={2022}
}

@article{cong2023attribute,
  title={Attribute-centric compositional text-to-image generation},
  author={Cong, Yuren and Min, Martin Renqiang and Li, Li Erran and Rosenhahn, Bodo and Yang, Michael Ying},
  journal={arXiv preprint arXiv:2301.01413},
  year={2023}
}

@inproceedings{yuan2023physdiff,
  title={Physdiff: Physics-guided human motion diffusion model},
  author={Yuan, Ye and Song, Jiaming and Iqbal, Umar and Vahdat, Arash and Kautz, Jan},
  booktitle=iccv,
  year={2023}
}

@inproceedings{
feng2023trainingfree,
title={Training-Free Structured Diffusion Guidance for Compositional Text-to-Image Synthesis},
author={Weixi Feng and Xuehai He and Tsu-Jui Fu and Varun Jampani and Arjun Reddy Akula and Pradyumna Narayana and Sugato Basu and Xin Eric Wang and William Yang Wang},
booktitle=iclr,
year={2023}
}

@article{gu2023mamba,
  title={Mamba: Linear-time sequence modeling with selective state spaces},
  author={Gu, Albert and Dao, Tri},
  journal={arXiv preprint arXiv:2312.00752},
  year={2023}
}

@inproceedings{zhang2025motion,
  title={Motion Mamba: Efficient and Long Sequence Motion Generation},
  author={Zhang, Zeyu and Liu, Akide and Reid, Ian and Hartley, Richard and Zhuang, Bohan and Tang, Hao},
  booktitle={European Conference on Computer Vision},
  pages={265--282},
  year={2025},
  organization={Springer}
}

@article{gao2024guess,
  title={Guess: Gradually enriching synthesis for text-driven human motion generation},
  author={Gao, Xuehao and Yang, Yang and Xie, Zhenyu and Du, Shaoyi and Sun, Zhongqian and Wu, Yang},
  journal={IEEE Transactions on Visualization and Computer Graphics},
  year={2024}
}

@inproceedings{zhai2023language,
  title={Language-guided human motion synthesis with atomic actions},
  author={Zhai, Yuanhao and Huang, Mingzhen and Luan, Tianyu and Dong, Lu and Nwogu, Ifeoma and Lyu, Siwei and Doermann, David and Yuan, Junsong},
  booktitle=acmmm,
  year={2023}
}

@inproceedings{zhou2023ude,
  title={Ude: A unified driving engine for human motion generation},
  author={Zhou, Zixiang and Wang, Baoyuan},
  booktitle=cvpr,
  year={2023}
}

@inproceedings{tevet2022motionclip,
    title={MotionCLIP: Exposing Human Motion Generation to CLIP Space},
    author={Tevet, Guy and Gordon, Brian and Hertz, Amir and Bermano, Amit H and Cohen-Or, Daniel},
    booktitle=eccv,
    year={2022}
    }

@inproceedings{petrovich21actor,
    title = {Action-Conditioned 3{D} Human Motion Synthesis with Transformer {VAE}},
    author = {Petrovich, Mathis and Black, Michael J. and Varol, G{\"u}l},
    booktitle = iccv,
    year = {2021}
}

@inproceedings{petrovich22temos,
  title     = {{TEMOS}: Generating diverse human motions from textual descriptions},
  author    = {Petrovich, Mathis and Black, Michael J. and Varol, Gul},
  booktitle = eccv,
  year      = {2022}
}

@inproceedings{ghosh2021synthesis,
  title={Synthesis of Compositional Animations from Textual Descriptions},
  author={Ghosh, Anindita and Cheema, Noshaba and Oguz, Cennet and Theobalt, Christian and Slusallek, Philipp},
  booktitle=iccv,
  year={2021}
}

@inproceedings{ahuja2019language2pose,
  title={Language2pose: Natural language grounded pose forecasting},
  author={Ahuja, Chaitanya and Morency, Louis-Philippe},
  booktitle=tdv,
  year={2019}
}

@InProceedings{attt2m,
    author    = {Zhong, Chongyang and Hu, Lei and Zhang, Zihao and Xia, Shihong},
    title     = {AttT2M: Text-Driven Human Motion Generation with Multi-Perspective Attention Mechanism},
    booktitle = iccv,
    year      = {2023}
}

@inproceedings{chuan2022tm2t,
  title={TM2T: Stochastic and Tokenized Modeling for the Reciprocal Generation of 3D Human Motions and Texts},
  author={Guo, Chuan and Zuo, Xinxin and Wang, Sen and Cheng, Li},
  booktitle=eccv,
  year={2022}
}

@article{zhang2022motiondiffuse,
  title={MotionDiffuse: Text-Driven Human Motion Generation with Diffusion Model},
  author={Zhang, Mingyuan and Cai, Zhongang and Pan, Liang and Hong, Fangzhou and Guo, Xinying and Yang, Lei and Liu, Ziwei},
  journal={arXiv},
  year={2022}
}

@inproceedings{guo2022generating,
  title={Generating Diverse and Natural 3D Human Motions From Text},
  author={Guo, Chuan and Zou, Shihao and Zuo, Xinxin and Wang, Sen and Ji, Wei and Li, Xingyu and Cheng, Li},
  booktitle=cvpr,
  year={2022}
}

@article{tevet2022MDM,
  title={Human Motion Diffusion Model},
  author={Tevet, Guy and Raab, Sigal and Gordon, Brian and Shafir, Yonatan and Bermano, Amit H and Cohen-Or, Daniel},
  journal={arXiv},
  year={2022}
}

@inproceedings{petrovich24stmc,
    title     = {Multi-Track Timeline Control for Text-Driven 3D Human Motion Generation},
    author    = {Petrovich, Mathis and Litany, Or and Iqbal, Umar and Black, Michael J. and Varol, G{\"u}l and Peng, Xue Bin and Rempe, Davis},
    booktitle = cvprw,
    year      = {2024}
}

@inproceedings{goel2024iterative,
  title={Iterative motion editing with natural language},
  author={Goel, Purvi and Wang, Kuan-Chieh and Liu, C Karen and Fatahalian, Kayvon},
  booktitle={ACM SIGGRAPH 2024 Conference Papers},
  year={2024}
}

@inproceedings{yang2023synthesizing,
  title={Synthesizing long-term human motions with diffusion models via coherent sampling},
  author={Yang, Zhao and Su, Bing and Wen, Ji-Rong},
  booktitle=acmmm,
  year={2023}
}

@inproceedings{xie2024towards,
  title={Towards Detailed Text-to-Motion Synthesis via Basic-to-Advanced Hierarchical Diffusion Model},
  author={Xie, Zhenyu and Wu, Yang and Gao, Xuehao and Sun, Zhongqian and Yang, Wei and Liang, Xiaodan},
  booktitle=aaai,
  year={2024}
}

@article{xie2023omnicontrol,
  title={Omnicontrol: Control any joint at any time for human motion generation},
  author={Xie, Yiming and Jampani, Varun and Zhong, Lei and Sun, Deqing and Jiang, Huaizu},
  journal={arXiv preprint arXiv:2310.08580},
  year={2023}
}

@inproceedings{wang2023fg,
  title={Fg-t2m: Fine-grained text-driven human motion generation via diffusion model},
  author={Wang, Yin and Leng, Zhiying and Li, Frederick WB and Wu, Shun-Cheng and Liang, Xiaohui},
  booktitle=iccv,
  year={2023}
}

@inproceedings{shafir2024priormdm,
  title={Human Motion Diffusion as a Generative Prior},
  author={Shafir, Yoni and Tevet, Guy and Kapon, Roy and Bermano, Amit Haim},
  year={2024},
  booktitle=iclr
}

@article{humantomato,
  title={HumanTOMATO: Text-aligned Whole-body Motion Generation},
  author={Lu, Shunlin and Chen, Ling-Hao and Zeng, Ailing and Lin, Jing and Zhang, Ruimao and Zhang, Lei and Shum, Heung-Yeung},
  journal={arxiv:2310.12978},
  year={2023}
}

@inproceedings{karunratanakul2023guided,
  title={Guided motion diffusion for controllable human motion synthesis},
  author={Karunratanakul, Korrawe and Preechakul, Konpat and Suwajanakorn, Supasorn and Tang, Siyu},
  booktitle=iccv,
  year={2023}
}

@article{ren2023realistic,
      title={Realistic Human Motion Generation with Cross-Diffusion Models},
      author={Ren, Zeping and Huang, Shaoli and Li, Xiu},
      journal={arXiv preprint arXiv:2312.10993},
      year={2023}
  }

@article{jin2024act,
  title={Act as you wish: Fine-grained control of motion diffusion model with hierarchical semantic graphs},
  author={Jin, Peng and Wu, Yang and Fan, Yanbo and Sun, Zhongqian and Yang, Wei and Yuan, Li},
  journal=nips,
  year={2023}
}

@inproceedings{han2024amd,
  title={AMD: Autoregressive Motion Diffusion},
  author={Han, Bo and Peng, Hao and Dong, Minjing and Ren, Yi and Shen, Yixuan and Xu, Chang},
  booktitle=aaai,
  year={2024}
}

@article{lou2023diversemotion,
  title={Diversemotion: Towards diverse human motion generation via discrete diffusion},
  author={Lou, Yunhong and Zhu, Linchao and Wang, Yaxiong and Wang, Xiaohan and Yang, Yi},
  journal={arXiv preprint arXiv:2309.01372},
  year={2023}
}

@inproceedings{kong2023priority,
  title={Priority-centric human motion generation in discrete latent space},
  author={Kong, Hanyang and Gong, Kehong and Lian, Dongze and Mi, Michael Bi and Wang, Xinchao},
  booktitle=cvpr,
  year={2023}
}

@inproceedings{TEACH:3DV:2022, 
  title = {{TEACH}: {T}emporal {A}ction {C}ompositions for {3D} {H}umans},
  author = {Athanasiou, Nikos and Petrovich, Mathis and Black, Michael J. and Varol, G{\"u}l},
  booktitle = {{International Conference on 3D Vision (3DV)}},
  year = {2022} 
  }

@inproceedings{rombach2022high,
  title={High-resolution image synthesis with latent diffusion models},
  author={Rombach, Robin and Blattmann, Andreas and Lorenz, Dominik and Esser, Patrick and Ommer, Bj{\"o}rn},
  booktitle=cvpr,
  year={2022}
}

@inproceedings{ho2020denoising,
  title={Denoising diffusion probabilistic models},
  author={Ho, Jonathan and Jain, Ajay and Abbeel, Pieter},
  booktitle=nips,
  year={2020}
}

@article{chen2022mld,
  author = {Xin, Chen and Jiang, Biao and Liu, Wen and Huang, Zilong and Fu, Bin and Chen, Tao and Yu, Jingyi and Yu, Gang},
  title = {Executing your Commands via Motion Diffusion in Latent Space},
  journal = {arXiv},
  year = {2022},
}

@inproceedings{SINC:ICCV:2023,
  author={Athanasiou, Nikos and Petrovich, Mathis and Black, Michael J. and Varol, G{\"u}l},
  title={{SINC}: Spatial Composition of {3D} Human Motions for Simultaneous Action Generation},
  booktitle=iccv,
  year = {2023}
}

@article{zhang2023motiongpt,
  title={MotionGPT: Finetuned LLMs are General-Purpose Motion Generators},
  author={Zhang, Yaqi and Huang, Di and Liu, Bin and Tang, Shixiang and Lu, Yan and Chen, Lu and Bai, Lei and Chu, Qi and Yu, Nenghai and Ouyang, Wanli},
  journal={arXiv},
  year={2023}
}
}

\clearpage
\appendix
\clearpage
\appendix
\vspace*{1em}{\centering\large\bf%
Appendix
\vspace*{1.5em}}

In this appendix, we present:
\begin{itemize}    
    \item Section~\ref{sec:supp_implementation}: Training and inference details of \textsc{DeMoGen}. \\

    \item Section~\ref{sec:supp_demogen_exp}: Ablations on the hyper-parameters of \textsc{DeMoGen-Exp} \\

    \item Section~\ref{sec:supp_demogen_oss}: Ablations on orthogonalization loss $\mathcal{L}_\text{Ortho}$ weight $\alpha_o$ for \textsc{DeMoGen-OSS}. \\

    \item Section~\ref{sec:supp_demogen_sc}: Ablations on orthogonalization loss $\mathcal{L}_\text{Ortho}$ weight $\alpha_o$ and semantic consistency loss $\mathcal{L}_\text{SC}$ weight $\alpha_{sc}$ for \textsc{DeMoGen-SC}. \\

    \item Section~\ref{sec:supp_additional_multi_compose_results}: Additional results of multi-concept motion generation. \\
    
    \item Section~\ref{sec:supp_nb_concepts}: Ablation studies on the number of decomposed motion concepts $K$. \\
    
    \item Section~\ref{sec:supp_t2m_decompml}: Text-to-motion evaluation on the extended HumanML3D, \ie, DeCompML.\\

    \item Section~\ref{sec:supp_other_models}: Additional results of our compositional training paradigm on MLD and MotionDiffuse \\

    \item Section~\ref{sec:supp_details_data_eval}: More details on datasets and evaluation metrics.\\

\end{itemize}

\begin{figure*}[t]
\centering
\begin{minipage}{0.7\textwidth}
\begin{algorithm}[H]
    \centering
    \caption{Text-to-Motion Generation and Compositional Motion Generation}
    \label{alg:supp_t2m_comp}
    \begin{algorithmic}[1]
        \STATE \textbf{Require}: Frozen VAE decoder $\mathcal{D}$, denoising network $\epsilon_{\theta}$, diffusion timestep $T$, text embeddings $\mC = \{\vc_k\}_{k=1}^K$, a latent feature $\vz_T \sim \mathcal{N}(0, 1)$
        \FOR{$t = T, \ldots, 1$}
            \IF{ mod is \textbf{latent-aware} }
            \STATE $\epsilon_{pred} \gets \textstyle{\sum_{k=1}^{K}} \epsilon_\theta(\vz_t, \vc_k, t)$ \\ 
            \ENDIF \\
            \IF{ mod is \textbf{semantic-aware} }
            \STATE $\epsilon_{pred} \gets \epsilon_\theta(\vz_t, \mC, t)$  \hspace{0.6cm} {\color{gray}\small{// \textit{Replace the cross-attention with DCA}}} \\
            \ENDIF \\
            \STATE {\color{gray}\small{// \textit{Run denoising step}}} \\
            \STATE $\vz_{t-1} = \frac{1}{\sqrt{\alpha}_t}(\vz_t -  \eta_t \epsilon_{pred}) + \mathcal{N}(0, \hat{\beta_t}I).$ 
            \STATE \hspace{\algorithmicindent} where $\alpha_t = 1 - \beta_t$, $\bar{\alpha}_t = \prod_{i=1}^t \alpha_i$, and $\eta_t = \frac{1-\alpha_t}{\sqrt{1-\bar{\alpha}_t}}$.
        \ENDFOR \\
    \end{algorithmic}
\end{algorithm}
\end{minipage}
\end{figure*}

\begin{figure*}[t]
\centering
\begin{minipage}{0.7\textwidth}
\begin{algorithm}[H]
    \centering
    \caption{Decompositional Motion Generation}
    \label{alg:supp_decomp}
    \begin{algorithmic}[2]
        \STATE \textbf{Require}: Frozen VAE decoder $\mathcal{D}$, denoising network $\epsilon_{\theta}$, diffusion timestep $T$, text embeddings $\mC = \{\vc_k\}_{k=1}^K$, a set of latent features $\{\vz_T^k\}_{k=1}^K$ with $\vz^k_T \sim \mathcal{N}(0, 1)$
        \FOR{$t = T, \ldots, 1$}
            \STATE {\color{gray}\small{// \textit{For both latent-aware and semantic-aware settings}}} \\
            \FOR{$k=1, \ldots, K$}
                \STATE $\epsilon^k_{pred} \gets \epsilon_\theta(\vz^k_t, \vc_k, t)$ \\
                \STATE {\color{gray}\small{// \textit{Run denoising step}}} \\
                \STATE $\vz^k_{t-1} = \frac{1}{\sqrt{\alpha}_t}(\vz^k_t -  \eta_t \epsilon^k_{pred}) + \mathcal{N}(0, \hat{\beta_t}I).$ 
                \STATE \hspace{\algorithmicindent} where $\alpha_t = 1 - \beta_t$, $\bar{\alpha}_t = \prod_{i=1}^t \alpha_i$, and $\eta_t = \frac{1-\alpha_t}{\sqrt{1-\bar{\alpha}_t}}$.
            \ENDFOR \\
        \ENDFOR \\
    \end{algorithmic}
\end{algorithm}
\end{minipage}
\end{figure*}

\section{Implementation Details}
\label{sec:supp_implementation}
\subsection{Training}
For motion VAE, we apply downsampling in both the temporal and spatial dimensions to obtain a latent feature $\vz \in \mathbb{R}^{L' \times N_j \times d_m'}$, where $L'$ is the latent feature length with a downsampling rate of 4. $N_j$ and $d_m'$ denote the number of joints and latent feature dimension, which are set to 7 and 32, respectively. The VAE is initialized with pretrained weights from~\cite{hong2025salad} for fair comparison. For the diffusion model, we use a 5-layer transformer with a dimension of 256. The learning rate is initialized at 0.0002 and subsequently reduced to 0.00002 following 50,000 training iterations.

On the extended HumanML3D dataset, \ie, DeCompML, we finetune the latent diffusion models of two recent state-of-the-art approaches, EnergyMoGen~\cite{zhang2025energymogen} and SALAD~\cite{hong2025salad}, for 100K iterations and 300 epochs. The learning rates are set to 0.00001 and 0.00002, respectively.
For the experiments in Section~\ref{sec:supp_other_models}, we apply our compositional training paradigm to two classic and representative methods: MLD~\cite{chen2022mld} and MotionDiffuse~\cite{zhang2022motiondiffuse}. We follow their original training configurations.

\begin{table*}[t]
    \centering
    \setlength{\tabcolsep}{8pt}
    \resizebox{0.85\linewidth}{!}{

    \begin{tabular}{c c c c c c c}
    \toprule
    \multirow{2}{*}{$\tau$}  & \multicolumn{3}{c}{R-Precision $\uparrow$} & \multirow{2}{*}{FID $\downarrow$} & \multirow{2}{*}{MM-Dist $\downarrow$} & \multirow{2}{*}{Diversity $\rightarrow$} \\

    \cline{2-4}
    ~ & Top-1 & Top-2 & Top-3 \\
        
    \midrule
    \rowcolor{gray!20}
    \multicolumn{7}{c}{\textit{\textbf{latent-aware}}} \\ \midrule

    0.0 & \et{0.532}{.005} & \et{0.726}{.004} & \et{0.820}{.003} & \et{0.255}{.012} & \et{2.926}{.012} & \etb{9.707}{.111} \\ 
    0.3 & \et{0.562}{.004} & \et{0.756}{.004} & \et{0.844}{.003} & \et{0.108}{.004} & \et{\underline{2.753}}{.017} & \et{\underline{9.766}}{.104} \\ 
    0.5 & \et{\underline{0.565}}{.004} & \et{\underline{0.757}}{.003} & \et{\underline{0.845}}{.003} & \et{\underline{0.095}}{.004} & \et{2.757}{.011} & \et{9.849}{.102} \\
    0.7 & \etb{0.569}{.004} & \etb{0.760}{.005} & \etb{0.850}{.004} & \etb{0.078}{.003} & \etb{2.708}{.012} & \et{9.774}{.099} \\
    \midrule
    \rowcolor{gray!20}
    \multicolumn{7}{c}{\textit{\textbf{semantic-aware}}} \\ \midrule
    0.0 & \et{0.564}{.003} & \et{0.756}{.004} & \et{0.845}{.002} & \et{0.297}{.016} & \et{2.764}{.012} & \etb{9.784}{.057} \\ 
    0.3 & \et{0.579}{.004} & \et{0.769}{.003} & \et{0.858}{.003} & \et{0.118}{.007} & \et{2.694}{.009} & \et{9.858}{.054} \\ 
    0.5 & \et{\underline{0.583}}{.002} & \et{\underline{0.774}}{.004} & \etb{0.865}{.002} & \etb{0.100}{.006} & \et{\underline{2.632}}{.006} & \et{\underline{9.807}}{.063} \\ 
    0.7 & \etb{0.586}{.005} & \etb{0.776}{.003} & \et{\underline{0.863}}{.002} & \et{\underline{0.116}}{.008} & \etb{2.623}{.008} & \et{9.873}{057} \\ 
    
    \bottomrule
    \end{tabular}
    }
    \caption{\textbf{Ablation of text replacement rate $\tau$ in \textsc{DeMoGen-Exp} for text-to-motion on the HumanML3D test set.} \textbf{Bold} and \underline{underlined} denote the best and second-best results, respectively.}
    \label{tab:supp_trr_humanml3d}
\end{table*}

\begin{table*}[t]
    \centering
    \setlength{\tabcolsep}{8pt}
    \resizebox{0.85\linewidth}{!}{

    \begin{tabular}{c c c c c c c}
    \toprule
    \multirow{2}{*}{$\tau$}  & \multicolumn{3}{c}{R-Precision $\uparrow$} & \multirow{2}{*}{FID $\downarrow$} & \multirow{2}{*}{MM-Dist $\downarrow$} & \multirow{2}{*}{Diversity $\rightarrow$} \\

    \cline{2-4}
    ~ & Top-1 & Top-2 & Top-3 \\
        
    \midrule
    \rowcolor{gray!20}
    \multicolumn{7}{c}{\textit{\textbf{latent-aware}}} \\ \midrule

    0.0 & \et{0.544}{.005} & \et{0.738}{.005} & \et{0.830}{.003} & \etb{0.089}{.006} & \et{2.850}{.010} & \etb{9.703}{.090} \\ 
    0.3 & \et{\underline{0.551}}{.005} & \et{\underline{0.743}}{.006} & \et{\underline{0.834}}{.004} & \et{0.109}{.007} & \et{\underline{2.817}}{.023} & \et{\underline{9.735}}{.104} \\ 
    0.5 & \et{0.550}{.004} & \et{\underline{0.743}}{.002} & \et{\underline{0.834}}{.003} & \et{\underline{0.093}}{.007} & \et{\underline{2.817}}{.015} & \et{9.803}{.105} \\
    0.7 & \etb{0.559}{.004} & \etb{0.754}{.004} & \etb{0.842}{.004} & \etb{0.089}{.005} & \etb{2.758}{.014} & \et{9.756}{.115} \\
    \midrule
    \rowcolor{gray!20}
    \multicolumn{7}{c}{\textit{\textbf{semantic-aware}}} \\ \midrule
    0.0 & \et{0.547}{.004} & \et{0.737}{.002} & \et{0.827}{.004} & \et{0.163}{.010} & \et{2.826}{.007} & \et{9.864}{.056} \\ 
    0.3 & \et{0.561}{.005} & \et{0.750}{.003} & \et{0.841}{.003} & \et{0.111}{.006} & \et{2.765}{.010} & \et{9.809}{.062} \\ 
    0.5 & \et{\underline{0.565}}{.004} & \et{\underline{0.757}}{.004} & \et{\underline{0.845}}{.003} & \etb{0.095}{.006} & \et{\underline{2.720}}{.012} & \etb{9.771}{.056} \\ 
    0.7 & \etb{0.567}{.004} & \etb{0.760}{.003} & \etb{0.846}{.002} & \et{\underline{0.102}}{.007} & \etb{2.719}{.009} & \et{\underline{9.786}}{.058} \\ 
    
    \bottomrule
    \end{tabular}
    }
    \caption{\textbf{Ablation of text replacement rate $\tau$ in \textsc{DeMoGen-Exp} for motion composition on DeCompML.}}
    \label{tab:supp_trr_decomp}
\end{table*}

\subsection{Inference}
During inference, we sample 50 diffusion steps to generate motion from texts. The pseudocode for text-to-motion generation (\textsc{DeMoGen-Exp, -OSS, -SC}) and compositional motion generation (\textsc{DeMoGen-Exp}) inference is illustrated in Algorithm~\ref{alg:supp_t2m_comp}. Note that \textsc{DeMoGen-Exp} adopts a text replacement rate~$\tau$ during training, enabling high-quality text-to-motion generation by duplicating the single textual description. The ablation study for $\tau$ is provided in Section~\ref{sec:supp_demogen_exp}. Furthermore, Algorithm~\ref{alg:supp_decomp} presents the pseudocode for decompositional motion generation, where decomposed concepts can be recombined using Algorithm~\ref{alg:supp_t2m_comp}.

\section{Ablations on \textsc{DeMoGen-Exp}}
\label{sec:supp_demogen_exp}
We conduct ablative experiments for \textsc{DeMoGen-Exp} on HumanML3D, DeCompML, and MTT. Text-to-motion results are provided in Table~\ref{tab:supp_trr_humanml3d}. As $\tau$ increases from 0.0 to 0.7, we observe that the model consistently shows improved R-Precision, FID, and MM-Dist. The results for compositional motion generation on the DeCompML dataset (in Table~\ref{tab:supp_trr_decomp}) demonstrate similar findings. $\tau=$0.7 achieves the best results under both latent-aware and semantic-aware settings. We also investigate the impact of $\tau$ on a more complicated benchmark, \ie, MTT. As shown in Table~\ref{tab:supp_trr_mtt}, we find that the latent-aware \textsc{DeMoGen-Exp} significantly outperforms the current state-of-the-art model, \ie, EnergyMoGen~\cite{zhang2025energymogen} across all settings of $\tau$. As for the semantic-aware model, we achieve improvements in most metrics, such as R-Precision, TMR-Score, and Transition distance, except for a decline in FID.

It is worth noting that the semantic-aware model performs better when the input conditions remain within the training distribution (HumanML3D and DeCompML). In contrast, the latent-aware model demonstrates stronger generalization in out-of-domain scenarios (MTT). Please note that all experimental results in Table~\ref{tab:supp_trr_humanml3d}–\ref{tab:supp_trr_mtt} are obtained using models trained on the HumanML3D dataset (with decomposed textual descriptions from DeCompML).

\begin{table*}
\centering
\setlength{\tabcolsep}{8pt}
\resizebox{0.8\linewidth}{!}{
\begin{tabular}{l|cc|cc|cc}
\toprule
\multirow{2}{*}{Methods} & \multicolumn{2}{c|}{R-Precision} & \multicolumn{2}{c|}{TMR-Score $\uparrow$} & \multirow{2}{*}{\text{FID} $\downarrow$} & Transition \\
& R@1 $\uparrow$ & R@3 $\uparrow$ & M2T & M2M & & distance $\downarrow$ \\
\midrule
\rowcolor{gray!20}
\multicolumn{7}{c}{\textit{\textbf{latent-aware}}} \\ \midrule
EnergyMoGen (latent) & 9.7  & 19.6 & 0.547 & 0.521 & 0.917 & \underline{1.6} \\ \midrule
\textsc{DeMoGen-Exp} ($\tau$=0.0) & \textbf{16.5}  & \textbf{31.9} & 0.594 & 0.566 & 0.630 & 1.7 \\
\textsc{DeMoGen-Exp} ($\tau$=0.3) & \underline{16.3}  & \underline{31.7} & 0.594 & \textbf{0.570} & \textbf{0.607} & \underline{1.6} \\
\textsc{DeMoGen-Exp} ($\tau$=0.5) & 15.3  & 31.1 & \underline{0.596} & 0.563 & 0.648 & \textbf{1.5} \\
\textsc{DeMoGen-Exp} ($\tau$=0.7) & 16.2  & \textbf{31.9} & \textbf{0.597} & \textbf{0.570} & \underline{0.621} & \underline{1.6} \\
\midrule
\rowcolor{gray!20}
\multicolumn{7}{c}{\textit{\textbf{semantic-aware}}} \\ \midrule
EnergyMoGen (semantic) & \underline{15.1} & 27.5 & \underline{0.585} & \underline{0.567} & \textbf{0.569} & 2.2 \\ \midrule
\textsc{DeMoGen-Exp} ($\tau$=0.0) & 13.5 & 27.1 & 0.577 & 0.560 & \underline{0.606} & \underline{2.1} \\
\textsc{DeMoGen-Exp} ($\tau$=0.3) & 14.1 & 28.2 & 0.581 & 0.563 & 0.648 & \textbf{1.8} \\
\textsc{DeMoGen-Exp} ($\tau$=0.5) & \textbf{15.2}  & \textbf{30.1} & \textbf{0.594} & 0.564 & 0.643 & \textbf{1.8} \\
\textsc{DeMoGen-Exp} ($\tau$=0.7) & 14.8 & \underline{29.3} & 0.584 & \textbf{0.568} & 0.631 & \textbf{1.8} \\
\bottomrule
\end{tabular}
}
\caption{\textbf{Quantitative comparison on the MTT~\cite{petrovich24stmc} dataset.} We compare our approach with EnergyMoGen and analyze the impact of the text replacement rate~$\tau$.}
\label{tab:supp_trr_mtt}
\end{table*}

\section{Ablations on \textsc{DeMoGen-OSS}}
\label{sec:supp_demogen_oss}
We study the impact of orthogonalization loss weight $\alpha_o$ in \textsc{DeMoGen-OSS}, and the results are reported in Table~\ref{tab:supp_alpha_o}. All tested values of $\alpha_o$ deliver competitive performance. Notably, $\alpha_o = 2.0$ yields the best performance for the latent-aware configuration, while $\alpha_o = 1.0$ performs best under the semantic-aware setting.

\begin{table*}[t]
    \centering
    \setlength{\tabcolsep}{8pt}
    \resizebox{0.9\linewidth}{!}{

    \begin{tabular}{c c c c c c c}
    \toprule
    \multirow{2}{*}{$\alpha_o$} & \multicolumn{3}{c}{R-Precision $\uparrow$} & \multirow{2}{*}{FID $\downarrow$} & \multirow{2}{*}{MM-Dist $\downarrow$} & \multirow{2}{*}{Diversity $\rightarrow$} \\
    \cline{2-4}
    ~ & Top-1 & Top-2 & Top-3 \\
    \midrule
    \rowcolor{gray!20}
    \multicolumn{7}{c}{\textit{\textbf{latent-aware}}} \\ \midrule
    0.0 & \et{0.582}{.005} & \et{0.771}{.003} & \et{0.855}{.003} & \et{0.138}{.005} & \et{2.642}{.008} & \et{9.851}{.147} \\ 
    0.1 & \et{\underline{0.584}}{.007} & \et{0.771}{.005} & \et{0.857}{.001} & \et{0.097}{.004} & \et{2.641}{.016} & \et{9.781}{.096} \\
    0.5 & \et{\underline{0.584}}{.003} & \et{\underline{0.775}}{.004} & \et{\underline{0.860}}{.003} & \et{\underline{0.092}}{.005} & \et{\underline{2.629}}{.009} & \etb{9.729}{.145} \\ 
    1.0 & \et{0.583}{.003} & \et{0.772}{.005} & \et{0.857}{.003} & \etb{0.074}{.004} & \et{2.635}{.009} & \et{9.799}{.107} \\ 
    2.0 & \etb{0.588}{.004} & \etb{0.778}{.002} & \etb{0.861}{.003} & \et{\underline{0.092}}{.003} & \etb{2.625}{.007} & \et{\underline{9.779}}{.120} \\ 
    \midrule
    \rowcolor{gray!20}
    \multicolumn{7}{c}{\textit{\textbf{semantic-aware}}} \\ \midrule
    0.0 & \et{\underline{0.583}}{.004} & \et{0.771}{.003} & \et{0.856}{.003} & \etb{0.093}{.005} & \et{2.649}{.009} & \etb{9.797}{.123} \\
    0.1 & \et{\underline{0.583}}{.005} & \etb{0.776}{.002} & \etb{0.861}{.003} & \et{0.113}{.005} & \et{\underline{2.648}}{.011} & \et{9.928}{.112} \\
    0.5 & \et{\underline{0.583}}{.003} & \etb{0.776}{.003} & \etb{0.861}{.002} & \et{0.112}{.005} & \et{2.662}{.010} & \et{9.929}{.126} \\
    1.0 & \etb{0.584}{.002} & \et{\underline{0.774}}{.003} & \et{\underline{0.858}}{.001} & \et{\underline{0.104}}{.005} & \etb{2.637}{.014} & \et{\underline{9.877}}{.127} \\
    2.0 & \et{0.578}{.004} & \et{0.769}{.003} & \et{0.855}{.004} & \et{0.107}{.004} & \et{2.665}{.010} & \et{9.905}{.114} \\
    \bottomrule
    \end{tabular}
    }
    \caption{\textbf{Ablation of orthogonalization loss weight $\alpha_o$ for \textsc{DeMoGen-OSS} on the HumanML3D test set.} We find that $\alpha_o=$2 and $\alpha_o=$1 yield the best performance for latent-aware and semantic-aware \textsc{DeMoGen-OSS}, respectively.}
    \label{tab:supp_alpha_o}
\end{table*}

\section{Ablations on Loss Weights $\alpha_{sc}$ and $\alpha_o$ for \textsc{DeMoGen-SC}}
\label{sec:supp_demogen_sc}
The results are provided in Table~\ref{tab:supp_alpha_o_sc}. We first conduct an ablation study on the scaling loss weight $\alpha_{sc}$, building on the findings presented in Section~\ref{sec:supp_demogen_oss}. We then further examine the performance of the model without orthogonalization loss. For the latent-aware \textsc{DeMoGen-SC}, although setting $\alpha_{\text{sc}}$ to 1 slightly degrades performance, we retain this value to encourage the model to learn from the decomposed text. Conversely, for the semantic-aware model, setting $\alpha_{\text{sc}}$ to 1 achieves the best performance. Meanwhile, the performance decreases when the orthogonalization loss is removed, further demonstrating its effectiveness.

\begin{table*}[t]
    \centering
    \setlength{\tabcolsep}{8pt}
    \resizebox{0.9\linewidth}{!}{

    \begin{tabular}{c c c c c c c c}
    \toprule
    \multirow{2}{*}{$\alpha_o$} & \multirow{2}{*}{$\alpha_{sc}$} & \multicolumn{3}{c}{R-Precision $\uparrow$} & \multirow{2}{*}{FID $\downarrow$} & \multirow{2}{*}{MM-Dist $\downarrow$} & \multirow{2}{*}{Diversity $\rightarrow$} \\
    \cline{3-5}
    ~ & ~ & Top-1 & Top-2 & Top-3 \\
    \midrule
    \rowcolor{gray!20}
    \multicolumn{8}{c}{\textit{\textbf{latent-aware}}} \\ \midrule
    2.0 & 0.1 & \etb{0.569}{.002} & \etb{0.761}{.004} & \etb{0.847}{.006} & \etb{0.112}{.006} & \etb{2.712}{.016} & \etb{9.763}{.125} \\
    2.0 & 0.5 & \et{0.560}{.004} & \et{0.756}{.003} & \et{\underline{0.846}}{.002} & \et{0.152}{.004} & \et{2.754}{.010} & \et{9.935}{.116} \\ 
    2.0 & 1.0 & \et{\underline{0.565}}{.003} & \et{\underline{0.757}}{.004} & \et{\underline{0.846}}{.003} & \et{\underline{0.121}}{.005} & \et{\underline{2.739}}{.009} & \et{\underline{9.933}}{.131} \\
    0.0 & 1.0 & \et{0.560}{.004} & \et{0.752}{.004} & \et{0.845}{.003} & \et{0.185}{.008} & \et{2.742}{.006} & \et{9.976}{.151} \\
    \midrule
    \rowcolor{gray!20}
    \multicolumn{8}{c}{\textit{\textbf{semantic-aware}}} \\ \midrule
    1.0 & 0.1 & \et{\underline{0.564}}{.003} & \et{\underline{0.755}}{.006} & \et{\underline{0.845}}{.005} & \et{\underline{0.132}}{.007} & \et{2.741}{.016} & \et{9.954}{.182} \\ 
    1.0 & 0.5 & \et{0.557}{.003} & \et{0.751}{.002} & \et{0.841}{.003} & \et{0.174}{.006} & \et{2.774}{.012} & \et{9.924}{.154} \\
    1.0 & 1.0 & \etb{0.565}{.002} & \etb{0.758}{.003} & \etb{0.846}{.003} & \et{0.138}{.005} & \etb{2.727}{.010} & \et{\underline{9.769}}{.158}  \\ 
    0.0 & 1.0 & \et{0.561}{.003} & \et{\underline{0.755}}{.004} & \et{0.844}{.003} & \etb{0.080}{.004} & \et{\underline{2.734}}{.009} & \etb{9.712}{.145} \\ 
    \bottomrule
    \end{tabular}
    }
    \caption{\textbf{Ablation of loss hyper-parameters in \textsc{DeMoGen-SC} on the HumanML3D test set.}}
    \label{tab:supp_alpha_o_sc}
\end{table*}

\section{Additional Results of Multi-concept Motion Generation}
\label{sec:supp_additional_multi_compose_results}
The results are shown in Table~\ref{tab:supp_multiconcept}. We observe that the semantic-aware setting surpasses the latent-aware setting on both \textsc{DeMoGen-OSS} and \textsc{DeMoGen-SC}, indicating its stronger ability to generate motions from a single complex textual description. Please refer to Table~\ref{tab:supp_trr_mtt} for the results and analysis of motion composition.

\begin{table*}
\centering
\setlength{\tabcolsep}{8pt}
\resizebox{0.8\linewidth}{!}{
\begin{tabular}{l|cc|cc|cc}
\toprule
\multirow{2}{*}{Methods} & \multicolumn{2}{c|}{R-Precision} & \multicolumn{2}{c|}{TMR-Score $\uparrow$} & \multirow{2}{*}{\text{FID} $\downarrow$} & Transition \\
& R@1 $\uparrow$ & R@3 $\uparrow$ & M2T & M2M & & distance $\downarrow$ \\

\midrule
\multicolumn{7}{c}{Multi-concept motion generation (single text)} \\
\midrule
\textsc{DeMoGen-OSS}(latent) & \underline{14.9} & \underline{29.5} & \underline{0.584} & \textbf{0.57} & \underline{0.580} & \underline{2.6} \\
\textsc{DeMoGen-OSS} (semantic) & \textbf{15.4} & 29.2 & \textbf{0.585} & \underline{0.571} & 0.598 & \textbf{2.4} \\ \midrule
\textsc{DeMoGen-SC} (latent) & 14.3 & \textbf{29.7} & 0.578 & 0.568 & 0.585 & 2.7 \\
\textsc{DeMoGen-SC} (semantic) & 14.8 & 29.1 & 0.578 & 0.569 & \textbf{0.575} & \underline{2.6} \\
\bottomrule        
\end{tabular}
}
\caption{\textbf{Quantitative results on MTT~\cite{petrovich24stmc}}. We present the multi-concept generation results of the semantic-aware \textsc{DeMoGen}. The metrics are computed following STMC~\cite{petrovich24stmc}.}
\label{tab:supp_multiconcept}
\end{table*}

\begin{table*}[t]
    \centering
    \setlength{\tabcolsep}{8pt}
    \resizebox{0.85\linewidth}{!}{

    \begin{tabular}{c c c c c c c}
    \toprule
    \multirow{2}{*}{$K$}  & \multicolumn{3}{c}{R-Precision $\uparrow$} & \multirow{2}{*}{FID $\downarrow$} & \multirow{2}{*}{MM-Dist $\downarrow$} & \multirow{2}{*}{Diversity $\rightarrow$} \\

    \cline{2-4}
    ~ & Top-1 & Top-2 & Top-3 \\
        
    \midrule
    \rowcolor{gray!20}
    \multicolumn{7}{c}{\textsc{DeMoGen-Exp}} \\ \midrule
    2 & \etb{0.569}{.004} & \etb{0.760}{.005} & \etb{0.850}{.004} & \etb{0.078}{.003} & \etb{2.708}{.012} & \et{\underline{9.774}}{.099} \\
    3 & \et{\underline{0.556}}{.005} & \et{\underline{0.749}}{.005} & \et{\underline{0.840}}{.003} & \et{\underline{0.110}}{.005} & \et{\underline{2.786}}{.016} & \et{9.822}{.133} \\ 
    4 & \et{0.553}{.005} & \et{0.748}{.004} & \et{0.836}{.002} & \et{0.157}{.005} & \et{2.795}{.011} & \etb{9.743}{.116} \\
    \midrule
    \rowcolor{gray!20}
    \multicolumn{7}{c}{\textsc{DeMoGen-OSS}} \\ \midrule
    2 & \etb{0.588}{.004} & \etb{0.778}{.002} & \et{\underline{0.861}}{.003} & \etb{0.092}{.003} & \etb{2.625}{.007} & \etb{9.779}{.120} \\
    4 & \et{\underline{0.584}}{.007} & \et{\underline{0.776}}{.002} & \etb{0.862}{.002} & \et{\underline{0.100}}{.006} & \et{\underline{2.636}}{.016} & \et{\underline{9.801}}{.247} \\
    
    \bottomrule
    \end{tabular}
    }
    \caption{\textbf{Ablation on the number of decomposed concepts $K$ on the HumanML3D test set.}}
    \label{tab:supp_k}
\end{table*}

\section{Ablation Studies on the Number of Decomposed Motion Concepts $K$}
\label{sec:supp_nb_concepts}
We explore the effect of the number of decomposed motion concepts $K$ on HumanML3D. GPT-4.1 is used to generate decomposed texts with three or four components (see Section~\ref{sec:supp_details_data_eval} for more details), and a latent-aware \textsc{DeMoGen-Exp} is trained accordingly. As shown in Table~\ref{tab:supp_k}, with an increasing number of decomposed motion concepts $K$, several key metrics, including FID and R-Precision, consistently decline on \textsc{DeMoGen-Exp}. We believe that the decrease is mainly due to the limited information in HumanML3D texts, which constrains the extraction of high-quality components under three- or four-part decompositions. To verify this, we further conduct experiments on \textsc{DeMoGen-OSS}, which does not require decomposed textual instructions. We find that $K=4$ achieves performance comparable to that of $K=2$. Furthermore, $K=2$ also enables a larger number of concept compositions through iterative aggregation of energy scores.

\section{Text-to-motion Evaluation on the DeCompML Dataset}
\label{sec:supp_t2m_decompml}
In Section~\ref{sec:exp_discuss}, we demonstrate that leveraging the decomposed motions as additional training data can improve text-to-motion performance. In this section, we provide complementary evaluation metrics and further validate our dataset by presenting results on a classical VQ-VAE–based approach, \ie, T2M-GPT~\cite{Zhang_2023_CVPR} (as shown in Figure~\ref{tab:supp_HumanML}). The improvement for SALAD is mainly reflected in the FID metric, which we attribute to its already strong performance in terms of R-precision and MM-Dist. Both T2M-GPT and EnergyMoGen show improvements in motion smoothness (FID) and text–motion consistency (R-precision and MM-Dist).

\begin{table*}[t]
    \centering
    \setlength{\tabcolsep}{8pt}
    \resizebox{0.94\linewidth}{!}{

    \begin{tabular}{l c c c c c c}
    \toprule
    \multirow{2}{*}{Method} & \multicolumn{3}{c}{R-Precision $\uparrow$} & \multirow{2}{*}{FID $\downarrow$} & \multirow{2}{*}{MM-Dist $\downarrow$} & \multirow{2}{*}{Diversity $\rightarrow$} \\

    \cline{2-4}
    ~ & Top-1 & Top-2 & Top-3 \\
        
    \midrule
    T2M-GPT~\cite{Zhang_2023_CVPR} & \et{0.491}{.003} & \et{0.680}{.003} & \et{0.775}{.002} & \et{0.116}{.004} & \et{3.118}{.011} & \et{9.761}{.081} \\
    T2M-GPT~\cite{Zhang_2023_CVPR}$^*$ & \et{0.497}{.004} & \et{0.685}{.002} & \et{0.780}{.003} & \et{0.102}{.008} & \et{3.017}{.007} & \et{9.748}{.074} \\ \midrule
    EnergyMoGen~\cite{zhang2025energymogen} & \et{0.523}{.003} & \et{0.715}{.002} & \et{0.815}{.002} & \et{0.188}{.006} & \et{2.915}{.007} & \et{9.488}{.099} \\
    EnergyMoGen~\cite{zhang2025energymogen}$^*$ & \et{0.526}{.004} & \et{0.718}{.002} & \et{0.818}{.003} & \et{0.147}{.004} & \et{2.884}{.009} & \et{9.392}{.084} \\ \midrule
    SALAD~\cite{hong2025salad} & \et{0.581}{.003} & \et{0.769}{.003} & \et{0.857}{.002} & \et{0.076}{.002} & \et{2.649}{.009} & \et{9.696}{.096} \\
    SALAD~\cite{hong2025salad}$^*$ & \et{0.580}{.003} & \et{0.769}{.003} & \et{0.857}{.003} & \et{0.060}{.005} & \et{2.651}{.009} & \et{9.379}{.149} \\
    
    \bottomrule
    \end{tabular}
    }
    \caption{\textbf{Quantitative comparison on the HumanML3D test set.} $*$ denotes that the model is finetuned on DeCompML.}
    \label{tab:supp_HumanML}
\end{table*}

\section{Additional Results of Our Compositional Training Paradigm on MLD and MotionDiffuse}
\label{sec:supp_other_models}
To comprehensively validate our approach, we apply the compositional training paradigm to MLD~\cite{chen2022mld} and MotionDiffuse~\cite{zhang2022motiondiffuse}, the results are presented in Table~\ref{tab:supp_mld_motiondiff}. We train both models based on orthogonal self-supervision (OSS) under a latent-aware setting. Notably, we find that MotionDiffuse achieves better performance when predicting $\vx_0$. The experimental results demonstrate that our approach serves as a general training paradigm, which can be seamlessly integrated with existing diffusion models.

\begin{table*}[t]
    \centering
    \setlength{\tabcolsep}{8pt}
    \resizebox{\linewidth}{!}{

    \begin{tabular}{l c c c c c c}
    \toprule
    \multirow{2}{*}{Method} & \multicolumn{3}{c}{R-Precision $\uparrow$} & \multirow{2}{*}{FID $\downarrow$} & \multirow{2}{*}{MM-Dist $\downarrow$} & \multirow{2}{*}{Diversity $\rightarrow$} \\

    \cline{2-4}
    ~ & Top-1 & Top-2 & Top-3 \\
        
    \midrule
    MLD~\cite{chen2022mld} & \et{0.481}{.003} & \et{0.673}{.003} & \et{0.772}{.002} & \et{0.473}{.013} & \et{3.196}{.010} & \et{9.724}{.082} \\
    \textsc{DeMoGen}-MLD~\cite{chen2022mld} & \et{0.490}{.003} & \et{0.680}{.004} & \et{0.776}{.003} & \et{0.288}{.005} & \et{3.124}{.011} & \et{9.785}{.115} \\ \midrule
    
    MotionDiffuse~\cite{zhang2022motiondiffuse} & \et{0.491}{.001} & \et{0.681}{.001} & \et{0.782}{.001} & \et{0.630}{.001} & \et{3.113}{.001} & \et{9.410}{.049} \\
    MotionDiffuse~\cite{zhang2022motiondiffuse}$^\S$ & \et{0.523}{.002} & \et{0.714}{.004} & \et{0.807}{.002} & \et{0.287}{.005} & \et{2.912}{.010} & \et{9.456}{.107} \\
    
    \textsc{DeMoGen}-MotionDiffuse~\cite{zhang2022motiondiffuse}$^\S$ & \et{0.527}{.002} & \et{0.719}{.003} & \et{0.820}{.004} & \et{0.141}{.006} & \et{2.908}{.009} & \et{9.628}{.096} \\
    \bottomrule
    \end{tabular}
    }
    \caption{\textbf{Quantitative results on the HumanML3D test set.} $\S$ indicates that we train MotionDiffuse using an $\vx_0$-prediction objective and apply DDIM for inference.}
    \label{tab:supp_mld_motiondiff}
\end{table*}

\section{More Details on Datasets and Evaluation Metrics}
\label{sec:supp_details_data_eval}
\subsection{DeCompML}
This section describes how decomposed text in DeCompML is generated. We utilize a large language model and design prompts to automatically perform the decomposition of the given text. The text prompt is shown in Table~\ref{tab:supp_prompt_text}. We experiment with multiple large language models, including GPT-4o, GPT-4.1, GPT-5, and GPT-OSS-120B (OpenAI’s 120B open-source model). GPT-OSS-120B and GPT-4o often fail to effectively decompose the text, frequently producing empty outputs. GPT-5 requires longer inference time and sometimes generates additional irrelevant content. Therefore, we ultimately select GPT-4.1 for our decomposition tasks.

\begin{table*}[tb]
    \centering
    \begin{tabular}{p{0.95\linewidth}}
        \toprule
        \textbf{Prompt:} \\
        I have a text description of a human motion. Your task is to split this description into exactly two separate sentences. \\ \\
        Each sentence should describe one distinct human motion. If the original text contains style, emotion, speed, or environment information, 
        you must preserve it in the corresponding sentences. No additional information should be added if it is absent in the original text. \\ \\
        Focus on accurately capturing the original meaning. The two sentences, when combined as sequential or simultaneous motions, should
        reproduce the meaning of the original description. \\ \\
        STRICT OUTPUT FORMAT (must follow exactly): \\
        - Return ONLY the two sentences joined by a single ``\#". \\
        - Both \texttt{<sentence1>} and \texttt{<sentence2>} must be non-empty and contain meaningful words (not just spaces or punctuation). \\
        - No labels, no explanations, no numbering, no quotes, no code fences, no extra spaces, and do not repeat the input. \\
        - The output must match the pattern: \texttt{<sentence1>\#<sentence2>} \\ \\
        Example: \\
        Input: \texttt{a person is walking forward while waving his left hand.} \\
        Output: \texttt{a person is walking forward\#a person is waving his left hand.} \\
        Now split the following sentence: {} \\
        \bottomrule
    \end{tabular}
    \caption{\textbf{Prompt for splitting human motion descriptions into two distinct sentences.}}
    \label{tab:supp_prompt_text}
\end{table*}

\subsection{Evaluation Metrics}
We evaluate text-driven human motion generation using the models and metrics from Guo~\etal~\cite{guo2022generating}. Motion quality is measured by FID, text-motion alignment by R-Precision and MM-Dist, and diversity by Diversity and Multimodality metrics. The feature sets of ground-truth and generated motions are denoted as $\vm$ and $\hat{\vm}$, respectively.

\paragraph{R-Precision.} For each motion sequence, we use 32 text descriptions (one ground-truth and 31 randomly selected mismatched) and rank them based on the Euclidean distance between motion and text embeddings. We report Top-1, Top-2, and Top-3 accuracy for motion-to-text retrieval.

\paragraph{FID.} The Fréchet Inception Distance (FID) serves as a key indicator of the realism and quality of synthetic motions. It calculates the statistical difference between the feature space of the generated samples ($\hat{\vm}$) and the real data ($\vm$):
\begin{equation}
    \text{FID} = ||\mu_{\hat{\vm}} - \mu_{\vm}||^2 + \text{TR}(\Sigma_{\hat{\vm}} + \Sigma_{\vm} - 2\sqrt{\Sigma_{\hat{\vm}}\Sigma_{\vm}}).
\end{equation}
Here, $\mu$ represents the feature set mean, and $\Sigma$ is the covariance matrix. A key property is that lower values of FID imply higher fidelity to the real data distribution. 

\paragraph{MM-Dist.} We compute the average Euclidean distance between each text embedding and the corresponding motion embedding generated from that text, providing a measure of alignment between text and motion features.

\paragraph{Diversity.} We quantify the variation of motions generated from different text descriptions in the test set by randomly sampling 300 motion pairs. For each pair, we compute the Euclidean distance between their feature representations, and define the Diversity metric as:

\begin{equation}
\text{Diversity} = \frac{1}{300} \sum_{i=1}^{300} \left\| \hat{\vm}_1^{(i)} - \hat{\vm}_2^{(i)} \right\|,
\label{formula:diversity}
\end{equation}

where $\hat{\vm}_1^{(i)}$ and $\hat{\vm}_2^{(i)}$ denote the motion features of the $i$-th pair.

\paragraph{Multimodality.} To evaluate the variation among motions generated from the same textual description, we adopt a procedure similar to Diversity. Following Guo~\etal~\cite{guo2022generating}, we generate multiple samples per text and partition them into two random subsets. The Multimodality metric is calculated as the average Euclidean distance between paired features from these subsets, following the same formulation as the Diversity metric.

Following STMC~\cite{petrovich24stmc} and EnergyMoGen~\cite{zhang2025energymogen}, we evaluate compositional motion generation using R-Precision, TMR-Score, FID, and Transition Distance. The TMR-Score quantifies motion-text alignment via the cosine similarity of embeddings derived from the TMR model~\cite{petrovich2023tmr}, similar to MM-Dist. Transition Distance is computed as the Euclidean distance between consecutive frames.

\end{document}